\documentclass[letterpaper]{article}
\usepackage{preprint}
\usepackage[hyphens]{url}
\usepackage{graphicx}
\usepackage{natbib}
\usepackage{caption}
\usepackage{algorithm}
\usepackage{algorithmic}
\usepackage{booktabs}
\usepackage{longtable}

\providecommand{\FloatBarrier}{}

\title{Do Tabular Foundation Models Still Need Feature Engineering?}

\author{Yifan Wu, Pinjun Dong, Jiran Tao, Binyan Jiang}

\affiliations{%
  Department of Data Science, The Hong Kong Polytechnic University, Hong Kong\\%
  \texttt{yat-fan.wu@connect.polyu.hk}\quad%
  \texttt{pinjun2000@gmail.com}\quad%
  \texttt{22117758r@connect.polyu.hk}\quad%
  \texttt{by.jiang@polyu.edu.hk}%
}

\begin{document}
	
	\maketitle
	
	\begin{abstract}
		Feature engineering has long been a cornerstone of tabular machine learning. Tabular foundation models (TFMs) are pretrained on a wide range of tabular datasets and applied via in-context learning. Their rise raises a natural question: does manual feature construction still matter as these models become more capable? To answer this, we perform a controlled study across several versions of two major TFM families, testing a wide range of existing feature engineering techniques on benchmark datasets from TabArena. We find a consistent pattern: feature engineering gains are concentrated in earlier model generations and become negligible for the strongest models. These results suggest that stronger TFMs depend less on explicitly engineered input representations. In a complementary experiment, however, adding in-context information from related datasets still improves performance. Our findings indicate a shift in the source of performance gains for stronger TFMs: re-representing existing inputs becomes less effective, while providing additional task-relevant context remains beneficial.
	\end{abstract}
	
	\section{Introduction}
	
	Feature engineering improves predictive performance by transforming raw variables into representations that better expose useful statistical structure \citep{zheng2018feature}. Implicit in this practice is the assumption that such structure should be encoded into the input representation before learning, rather than inferred by the estimator itself.
	
	Tabular foundation models (TFMs) challenge this assumption. Pre-trained across many tabular tasks and applied through in-context learning, TFMs may reduce the need for task-specific feature engineering. As these models become more capable, statistical structure that previously required explicit transformations may become increasingly accessible from the original input. This raises a capability dependent question: does the marginal value of feature engineering decline as TFMs become stronger?
	
	Recent work has shown that feature-engineering pipelines can improve TFMs by explicitly exposing useful structure in the data \citep{tschalzev2026tabprep}. However, it remains unclear whether these gains reflect a persistent limitation of TFMs in recovering structure from raw inputs, or whether they are concentrated in particular models and datasets and weaken as TFMs become more capable. Existing evidence does not distinguish between these possibilities, as the value of feature engineering has not yet been systematically compared across model generations under controlled, within-estimator settings. 
	
	Answering this question requires evaluating feature engineering within each fixed estimator and then comparing its effect across successive TFM generations. We therefore study whether engineered representations provide consistent gains across model versions, datasets, and tasks. We further examine whether TFMs that benefit little from feature transformations can still exploit additional examples from related datasets. Together, these analyses investigate whether stronger TFMs become broadly less responsive to additional intervention, or whether the value of intervention shifts from re-representing existing inputs to providing additional task-relevant context.
	
	To carry out this study, we evaluate successive versions of the TabPFN and TabICL families on datasets from TabArena \citep{erickson2025tabarena}. For each dataset, split, and estimator, we use the model default configuration and native preprocessing while keeping the evaluation protocol fixed. This paired comparison isolates how the effect of feature engineering changes across model versions. In a separate experiment, we examine whether stronger TFMs can still benefit from additional source examples selected for compatibility with the target task.
	
	Across the evaluated TabArena tasks, generic feature engineering is not a reliable source of improvement for TFMs. Most transformations are neutral or harmful, while positive gains are concentrated in particular datasets and earlier model versions. For the strongest models, across dataset gains approach zero. These models nevertheless remain responsive to additional examples from related datasets. This contrast suggests that the declining value of generic feature transformations should not be interpreted as a general inability to exploit additional task-relevant information.
	
	Our contribution is a controlled empirical reassessment of feature engineering for TFMs. By isolating within-estimator gains across successive TabPFN and TabICL versions, we show that the benefits of generic feature engineering are concentrated in earlier models and a small subset of datasets, while becoming negligible for the strongest models. Our complementary context-augmentation experiment provides evidence that these models can still exploit additional task-relevant information, suggesting that the opportunity for improvement is shifting from re-representing existing inputs toward enriching the available task context.
	
	\section{Related Work}
	Feature engineering transforms input variables to expose
	structures that may be easier for a particular estimator to use.
	Common methods include numerical transformations, categorical
	encodings, discretization, and interaction construction. Prior
	work shows that engineered features interact with the estimator:
	different model classes can respond differently to the same
	engineered structure \citep{heaton2016empirical}. Tree-based and
	neural tabular models also differ in robustness to uninformative
	features, feature orientation, and irregular target functions
	\citep{grinsztajn2022tree}. Within neural models, transformations
	such as scaling and ranking can smooth irregular targets, but may
	also discard information or impair optimization, making their
	utility dependent on the feature and model setting
	\citep{beyazit2023inductive}. These findings motivate evaluating
	feature transformations separately for each estimator.
	
	Tabular foundation models introduce a different setting. TabPFN and
	TabICL are pretrained across tabular tasks and adapt to new datasets
	through in-context learning, while successive versions differ in
	architecture, pretraining, context capacity, and the dataset sizes
	they can accommodate
	\citep{hollmann2023tabpfn,hollmann2025tabpfn,
		grinsztajn2025tabpfn25,grinsztajn2026tabpfn3,
		qu2025tabicl,qu2026tabiclv2}. Such changes may alter which
	structures can be recovered directly from raw inputs and which still
	benefit from explicit transformations. Evidence obtained from one
	model version therefore does not establish whether the effect of
	feature engineering persists across later versions.
	
	TabPrep is closely related to our study
	\citep{tschalzev2026tabprep}. It generates feature variants targeting
	numerical interactions, group-conditional effects, and
	pseudo-categorical numerical structure, and evaluates them within the
	TabArena training, tuning, and ensembling pipeline. Its results show
	that engineered variants can improve performance within this broader
	pipeline, particularly when applied to TabPFN-2.5. Our study asks a
	complementary question: how does the paired effect of feature
	engineering, measured within each estimator under a common evaluation
	protocol, change across successive versions of TabPFN and TabICL?
	\section{Problem Setup and Estimand}

	\subsection{Notation}

Let the $d$-th tabular prediction task be
	\[
	\mathcal D_d=(X_d,y_d),
	\]
	where $X_d$ denotes the original predictors and $y_d$ denotes the
	classification or regression target. Let $m$ denote a fixed tabular
	foundation model estimator, including its checkpoint, default
	configuration, ensemble procedure, and native preprocessing pipeline.
	
	Following the standard view of feature engineering as an explicit
	transformation of the input representation before prediction
	\citep{zheng2018feature}, we represent each external
	feature engineering condition by a mapping
	\[
	T_k:\mathcal X_d\rightarrow\mathcal Z_{d,k}.
	\]
	The identity condition is $T_0(X)=X$, corresponding to no additional
	external feature engineering beyond the foundation model native
	preprocessing. Each $k>0$ denotes one pre-specified intervention, and
	$\mathcal K(d)$ denotes the set of non-identity interventions
	applicable to dataset $d$. Any data-dependent component of $T_k$ is
	fitted using only the corresponding training split.
	
	Let $\mu_{d,s,m,k}$ denote the test metric obtained on dataset $d$,
	evaluation split $s$, estimator $m$, and representation condition $k$.
	Let $\sigma_q\in\{-1,+1\}$ denote the orientation of metric $q$, with
	$\sigma_q=+1$ when larger values are better and $\sigma_q=-1$ when
smaller values are better.

\subsection{Feature engineering gain}

We define the direction normalized paired
gain over the identity representation as
	\[
	\delta_{d,s,m,k}
	=
	\sigma_q
	\left(
	\mu_{d,s,m,k}-\mu_{d,s,m,0}
	\right).
	\]
	Thus, $\delta_{d,s,m,k}>0$ consistently indicates that intervention
	$k$ outperforms the identity representation under the same split and
	estimator configuration.
	
	We summarize the repeated paired evaluations within each dataset by
	\[
	\Delta_{d,m,k}
	=
	\mathrm{median}_{s}
	\;
	\delta_{d,s,m,k}.
	\]
	Evaluation splits are treated as repeated measurements within a
	dataset, while datasets form the units of benchmark-level analysis.
	
	To determine whether any evaluated intervention improves over
	identity, we define the best observed feature-engineering gain as
	\[
	B_{d,m}
	=
	\max_{k\in\mathcal K(d)}
	\Delta_{d,m,k}.
	\]
	Accordingly, $B_{d,m}\leq0$ indicates that none of the evaluated
	interventions applicable to dataset $d$ achieves a positive median
	paired gain, and $B_{d,m}$ is an oracle style observation, which reports
	the best observed condition without implying that this condition
	could be selected without a validation set.

	\subsection{Ordered hypothesis tests.}
	For each version family, let \(D\) denote the number of benchmark datasets and \(G\) the number of estimator versions, ordered chronologically and indexed by \(g=1,\ldots,G\). We use \(B_{d,g}\) for the best-observed gain introduced above, evaluated on dataset \(d\) at version \(g\). The candidate set \(\mathcal K(d)\) is held fixed across versions, and datasets are the independent units of inference.

	For the TabPFN version families with \(G>2\), we use a one-sided exact
	Page test for an ordered decline in best-observed headroom
	\citep{page1963ordered}. Under the null, version labels are exchangeable
	within each dataset. The alternative is that larger \(B_{d,g}\) values
	tend to occur at earlier version positions.
	
	For each dataset \(d\), let \(r_{d,g}\) denote the ascending midrank of
	\(B_{d,g}\) across the \(G\) versions. The Page statistic is
	\[
	L
	=
	\sum_{d=1}^{D}\sum_{g=1}^{G}(G+1-g)r_{d,g}.
	\]
	Since earlier versions receive larger weights, larger values of \(L\)
	provide stronger evidence of an ordered decline. We compute the exact
	one-sided raw \(p\)-value by enumerating all within dataset permutations
	of the version labels.

	TabICL classification includes only two versions, so the ordered comparison
	reduces to a paired test. For each dataset, let
	\[
	C_d=B_{d,\mathrm{v1}}-B_{d,\mathrm{v2}},
	\]
	where \(C_d>0\) indicates a reduction in best-observed headroom from v1 to
	v2. We use
	\[
	T=\frac{1}{D}\sum_{d=1}^{D}C_d
	\]
	as the statistic in a one-sided exact sign-flip test. Under the null, the
	paired differences are symmetric about zero. The exact raw \(p\)-value is
	computed over all \(2^D\) sign assignments.

	The exact one-sided \(p\)-values from the three prespecified primary tests
	are adjusted jointly using the Benjamini--Hochberg procedure, with
	\(q<0.05\) as the primary multiplicity-adjusted significance criterion
	\citep{benjamini1995fdr}. We additionally report Holm-adjusted
	\(p\)-values to assess robustness under family-wise error control
	\citep{holm1979sequentially}. As descriptive effect summaries, we report
	the median paired first-to-last difference
	\[
	\widetilde C
	=
	\mathrm{median}_{d}\!\left(B_{d,1}-B_{d,G}\right)
	\]
	and
	\[
	N_{\downarrow}
	=
	\sum_{d=1}^{D}
	\mathbf{1}\!\left\{B_{d,1}>B_{d,G}\right\},
	\]
	is the number of datasets with a positive first-to-last difference.
	
	\section{Experiments}
	
	We evaluate 12 estimator versions from the TabPFN and TabICL families on 13 TabArena datasets (Table~\ref{tab:datasets}): eight classification and five regression tasks. The classification panel includes TabPFN v1, v2, v2.5, v2.6, v3, TabICL v1, and TabICL v2; all selected classification datasets satisfy the sample and feature limits of TabPFN v1 (at most 1,024 training samples and 100 features), ensuring that every generation is evaluated on the same datasets. This restriction is deliberate: small data tasks are the intended operating regime of TFMs and provide a stringent setting for testing whether feature engineering still supplies useful inductive bias when task specific evidence is limited. Regression experiments include TabPFN v2, v2.5, v2.6, v3, and TabICL v2, as both TabPFN v1 and TabICL v1 are classification only. All experiments were run on a Ubuntu server with 32 GB NVIDIA Tesla V100 and AMD Ryzen 9700X processor.
	
	\begin{table}[t]
		\centering
		\begin{tabular}{@{}lrcc@{}}
			\toprule
			Dataset & Samples & Features & Classes \\
			\midrule
			\multicolumn{4}{@{}l}{\textit{Classification}} \\
			Blood transfusion & 748 & 4 & 2 \\
			Diabetes & 768 & 8 & 2 \\
			Anneal & 898 & 38 & 5 \\
			Credit-g & 1,000 & 20 & 2 \\
			Maternal health & 1,014 & 6 & 3 \\
			QSAR biodeg & 1,054 & 41 & 2 \\
			Website phishing & 1,353 & 9 & 3 \\
			Fitness club & 1,500 & 6 & 2 \\
			\midrule
			\multicolumn{4}{@{}l}{\textit{Regression}} \\
			QSAR fish toxicity & 907 & 6 & --- \\
			Concrete strength & 1,030 & 8 & --- \\
			Healthcare insurance & 1,338 & 6 & --- \\
			Airfoil noise & 1,503 & 5 & --- \\
			Used Fiat 500 & 1,538 & 7 & --- \\
			\bottomrule
		\end{tabular}
		\caption{The 13 TabArena datasets used in this study, ordered by sample count within each task.}
		\label{tab:datasets}
	\end{table}
	
	For each dataset and estimator, we compare the identity baseline against 26 feature engineering conditions drawn from seven families: scale and power transforms, categorical encodings, discretization, interaction features, selection methods, the TabPrep composite pipeline, and two prompt-inspired conditions adapted from in-context learning. Each condition is a single atomic transformation; depending on column types, between 16 and 26 of these conditions apply to a given dataset. Data dependent components are fitted on the corresponding training split only.
	
	Every condition is evaluated on the 30 official TabArena splits (10 repeats $\times$ 3 folds). For each dataset, estimator, and feature engineering combination, we compute the paired gain against identity on the same split and summarize it by the within cell median \(\Delta_{d,m,k}\); the best observed gain \(B_{d,m}\) is the maximum of these medians over applicable non-identity conditions. Following the TabArena protocol~\citep{erickson2025tabarena}, we adopt macro-F1 for classification and sign reversed RMSE for regression, so that positive values consistently favor feature engineering.
	
	\begin{table*}[t]
		\centering
		\begin{tabular}{@{}p{3.2cm}cp{8.4cm}p{4.0cm}@{}}
			\toprule
			Family & Count & Methods & Role \\
			\midrule
			Numeric transforms & 5 & Standard; log; sqrt; Box--Cox; min--max & Transform numeric inputs \\
			\mbox{Categorical encoding} & 5 & One-hot; ordinal; frequency; target; crosses & Encode categorical inputs \\
			Binning & 2 & Equal-width; equal-frequency & Model threshold effects \\
			Interactions & 2 & Numeric interactions; group aggregation & Add higher-order structure \\
			Feature selection & 4 & Variance filter; collinearity filter; PCA; supervised selection & Remove redundant features \\
			TabPrep & 6 & Full pipeline; no-groupby; no-RSFC; no-arithmetic; no-categorical-interaction; no-OOF-target-encoding & Combine feature transformations \\
			Prompt-inspired & 2 & Shared-context kNN retrieval; three-fold OOF prediction feature & Alter context/model input \\
			\bottomrule
		\end{tabular}
		\caption{The 26 feature engineering conditions grouped by family. Applicability depends on dataset column types; the identity baseline is not shown.}
		\label{tab:fe}
	\end{table*}
	
	The prompt-inspired family adapts two forms of context augmentation from LLMs. The first constructs a shared ICL context by retrieving training rows most similar to the test fold~\citep{thomas2024retrieval}. The second is a tabular analogue of generated knowledge prompting: predictions from the same estimator are added as auxiliary inputs to its final prediction~\citep{liu2022generated}. These features are constructed using leakage safe three-fold OOF predictions, making the implementation a form of stacked generalization~\citep{wolpert1992stacked}.
	
	As an additional experiment, we examined whether providing task relevant context benefits the strongest TFMs. Among the possible forms of context augmentation, we considered a simple setting in which samples from pre-specified related datasets are added to the model in-context examples. Because related source datasets are typically distribution shifted
	relative to the target task, augmenting the context with the full source
	pool can introduce incompatible examples. We therefore introduce
	Nearest-Neighbor Reference Consistency (NNRC) screening to identify a
	target compatible source subset. NNRC first filters source samples by
	proximity to the target covariate distribution and then ranks the
	remaining samples by label consistency with a reference model fitted
	only on the target training data. The complete procedure is given in
	Algorithm~\ref{alg:nnrc}.
	
	\begin{algorithm}[t]
		\begingroup
		\small
		\caption{NNRC Screening}
		\label{alg:nnrc}
		\begin{algorithmic}[1]
			\REQUIRE Source pool $\mathcal S=(X^{\mathrm{src}},y^{\mathrm{src}})$; target training split $\mathcal D^{\mathrm{tr}}=(X^{\mathrm{tr}},y^{\mathrm{tr}})$; target test covariates $X^{\mathrm{te}}$; retention ratio $\lambda\in(0,1]$
			\ENSURE Selected source subset $\mathcal S^\star$
			\STATE \textbf{Stage 1: proximity screening (covariate space).}
			\STATE \quad Align the source and target schemas; impute each partition separately.
			\STATE \quad For each source sample $x$, compute its mean Gower distance $d(x)$ to its $k$ nearest neighbors in $X^{\mathrm{te}}$.
			\STATE \quad Fit a two-component Gaussian mixture on the scores $\{d(x)\}$ and keep the lower-mean component as the candidate set $\mathcal C$.
			\STATE \textbf{Stage 2: reference-consistency selection.}
			\STATE \quad Fit a reference model $f$ on $\mathcal D^{\mathrm{tr}}$.
			\STATE \quad For each $(x_i,y_i)\in\mathcal C$, compute the reference loss
			\STATE \qquad $\ell_i=\left\{\begin{array}{ll} |y_i-f(x_i)|, & \mathrm{regression}\\ -\log P_f(y_i\mid x_i), & \mathrm{classification}.\end{array}\right.$
			\STATE \quad Set the budget $m \gets \lceil \lambda |\mathcal C| \rceil$.
			\IF{regression}
			\STATE $\mathcal S^\star \gets$ the $m$ lowest loss candidates in $\mathcal C$.
			\ELSE
			\STATE Allocate per-class quotas $q_c$ proportional to the $y^{\mathrm{tr}}$ class frequencies via the largest remainder method, with $\sum_c q_c=m$.
			\STATE Within each class $c$, select the $q_c$ lowest loss candidates into $\mathcal S^\star$.
			\STATE Fill any remaining slots with the lowest loss unselected candidates.
			\ENDIF
			\RETURN $\mathcal S^\star$
		\end{algorithmic}
		\endgroup
	\end{algorithm}
	
	We restrict the context augmentation study to three target benchmarks:
	maternal health, concrete strength, and airfoil noise. For each task, we
	identified a publicly available, task relevant source dataset whose
	predictors could be semantically aligned with the complete target
	predictor set. No comparably matched public source was available for the
	remaining TabArena tasks. The analysis is therefore a feasibility study
	conditioned on source availability and schema compatibility, rather than
	a benchmark wide evaluation. The three source--target pairs exhibit different degrees of residual compatibility after predictor alignment. We characterize this difficulty qualitatively from the correspondence between their data-generating processes, the preservation of predictor and target semantics, and the extent of population or label shift. These characterizations are based on dataset provenance rather than downstream augmentation results. Table~\ref{tab:source-pairs} summarizes the resulting hierarchy.
	
	\begin{table*}[t]
		\centering
		\small
		\setlength{\tabcolsep}{1mm}
		\begin{tabular}{@{}lp{1.55in}p{1.3in}p{1.7in}l@{}}
			\toprule
			\textbf{Benchmark} & \textbf{Basis for alignment} & \textbf{Source data} & \textbf{Residual mismatch} & \textbf{Difficulty} \\
			\midrule
			Airfoil noise & {\raggedright Both include NACA 0012 aerofoil measurements and represent closely related aeroacoustic prediction problems~\citep{brooks1989airfoil,branch2025aerofoil}.\par} & {\raggedright A separate University of Bristol aeroacoustic and PIV campaign covering pre-stall, stall, and post-stall conditions.\par} & {\raggedright Differences in facility, operating conditions, flow regime, and measurement protocol.\par} & Lower \\
			Concrete strength & {\raggedright Both relate mixture composition and curing conditions to compressive strength~\citep{yeh1998concrete,martinez2025geopolymer}.\par} & {\raggedright A literature-assembled collection of geopolymer-concrete mixtures.\par} & {\raggedright Differences in binder chemistry, precursor and activator composition, curing protocols, and cross-study curation.\par} & Moderate \\
			Maternal health & {\raggedright Six physiological predictors can be aligned between the two Bangladesh maternal-health datasets~\citep{ahmed2020maternal,mojumdar2025maternal}.\par} & {\raggedright A later rural clinical dataset containing additional historical and physiological variables.\par} & {\raggedright Differences in patient population and label construction; the source labels missing the target mid-risk class.\par} & Higher \\
			\bottomrule
		\end{tabular}
		\caption{Source–target compatibility in context augmentation. Difficulty is residual mismatch after predictor alignment.}
		\label{tab:source-pairs}
	\end{table*}
	
	\section{Results}
	
	\subsection{Feature engineering gains are smaller in later model versions.}
	We distinguish the typical intervention effect, defined as the median gain across applicable interventions within each dataset, from the best observed effect, defined as the largest median paired gain among those interventions. The latter is an oracle summary of available headroom rather than the performance of a deployable selection procedure. Across datasets, the median best observed gain decreases from 1.21 to 0.25 macro-F1 percentage points between TabPFN v1 and v3 classification, from 2.06\% to 0.19\% relative RMSE reduction between TabPFN v2 and v3 regression, and from 0.73 to 0.10 macro-F1 points between TabICL v1 and v2 classification. In contrast, the typical intervention effect remains non-positive throughout.

	Exact ordered alternative tests support an aggregate decline in best observed feature engineering headroom (Table~\ref{tab:ordered-tests}). Page tests provide evidence for the ordered decline in TabPFN classification ($L=389$, one-sided $p=0.0174$) and regression ($L=139$, $p=0.0167$). For TabICL classification, the exact paired sign-flip test likewise supports lower headroom in v2 than in v1 ($T=0.412$ macro-F1 percentage points, $p=0.0234$). All three comparisons remain significant after Benjamini--Hochberg false-discovery-rate correction ($q=0.0234$); the more conservative Holm-adjusted value is borderline and lies just above 0.05 ($p_{\mathrm{Holm}}=0.0502$). The paired median first-to-last contraction is 1.095 macro-F1 percentage points for TabPFN classification, 1.590 relative-RMSE percentage points for TabPFN regression, and 0.429 macro-F1 percentage points for TabICL classification. Headroom declines from the first to the last evaluated version on 5/8, 4/5, and 6/8 datasets, respectively. Thus, the evidence concerns an aggregate ordered trend.

	\begin{figure*}[t]
		\centering
		\includegraphics[width=0.765\textwidth]{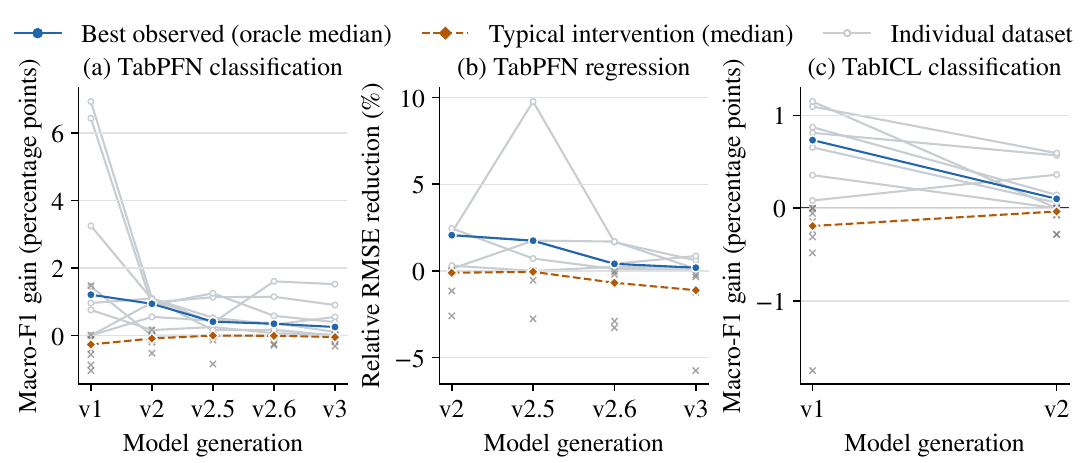}
		\caption{Feature engineering response across model generations. Per dataset and version, gains are medians over 30 paired splits; gray crosses and hollow circles show their median and maximum across interventions. Orange and blue marks are the respective across-dataset medians. Positive favors feature engineering; the maximum is oracle, not validation-selected. Units are macro-F1 points (classification) and relative RMSE reduction (regression).}
		\label{fig:generational-response}
	\end{figure*}

	\begin{figure*}[t]
		\centering
		\includegraphics[width=0.8\textwidth]{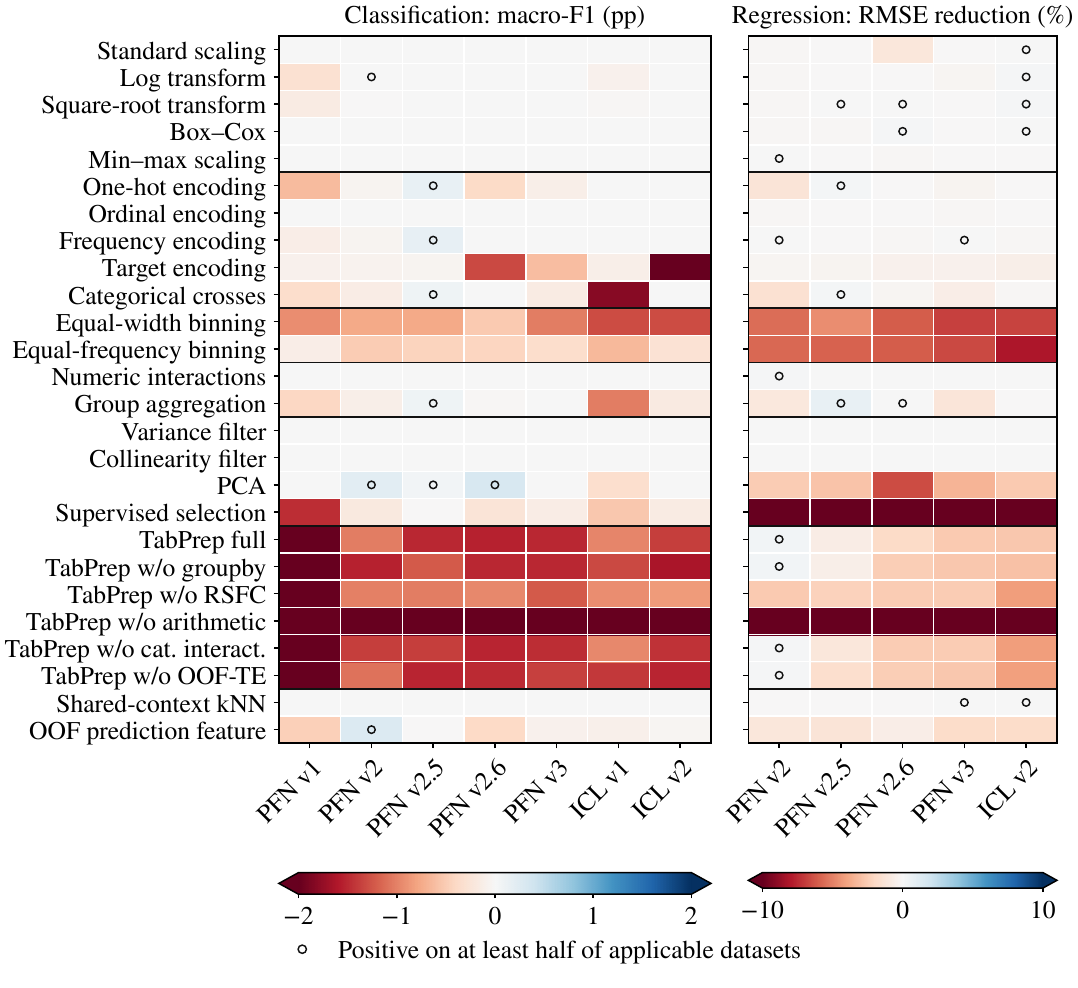}
		\caption{Across-dataset median effects for all 26 interventions and 12 estimators; positive favors feature engineering. Classification uses macro-F1 points (color limits $\pm2$); regression uses relative RMSE reduction ($\pm10\%$). Open circles mark positive effects on at least half of applicable datasets; rules separate intervention families.}
		\label{fig:intervention-heatmap}
	\end{figure*}
	
	\begin{table*}[t]
		\centering
		\small
		\setlength{\tabcolsep}{1mm}
		\begin{tabular}{@{}lcccc@{}}
			\toprule
			& \multicolumn{2}{c}{Classification (F1 pp)} & \multicolumn{2}{c}{Regression (relative RMSE \%)} \\
			Family & PFN v3 & ICL v2 & PFN v3 & ICL v2 \\
			\midrule
			Numeric transform & $+0.00$ (9/35) & $+0.00$ (1/35) & $-0.04$ (6/25) & $+0.00$ (18/25) \\
			Categorical encoding & $+0.00$ (2/25) & $+0.00$ (4/25) & $-0.23$ (5/13) & $-0.10$ (3/13) \\
			Binning & $-0.88$ (1/14) & $-0.96$ (2/14) & $-6.70$ (0/10) & $-7.40$ (0/10) \\
			Interactions & $+0.00$ (1/11) & $+0.00$ (0/11) & $-0.55$ (2/8) & $-0.01$ (3/8) \\
			Selection & $+0.00$ (5/31) & $+0.00$ (6/31) & $-0.93$ (0/20) & $-0.83$ (0/20) \\
			TabPrep & $-1.47$ (11/48) & $-1.58$ (4/48) & $-2.62$ (1/30) & $-4.13$ (0/30) \\
			Prompt-inspired & $+0.00$ (3/16) & $+0.00$ (2/16) & $-0.04$ (5/10) & $+0.00$ (4/10) \\
			\bottomrule
		\end{tabular}
		\caption{Feature engineering family effects for the strongest evaluated model variants. Each cell reports the median direction normalized effect across applicable (dataset, intervention) cells, followed by the number of positive cells over the number applicable. Positive values favor feature engineering.}
		\label{tab:strongest-family-summary}
	\end{table*}
	
	\begin{table}[t]
		\centering
		\small
		\setlength{\tabcolsep}{1mm}
		\begin{tabular}{@{}p{0.90in}cp{1.65in}@{}}
			\toprule
			Panel & $D/G$ & Exact test; statistic; raw $p$; median contraction (declining) \\
			\midrule
			TabPFN classification & 8/5 & Page; $L=389$; 0.0174; 1.095 (5/8) \\
			TabPFN regression & 5/4 & Page; $L=139$; 0.0167; 1.590 (4/5) \\
			TabICL classification & 8/2 & Sign-flip; $T=0.412$; 0.0234; 0.429 (6/8) \\
			\bottomrule
		\end{tabular}
		\caption{Exact dataset-level tests for declining best observed feature engineering headroom. Raw \(p\) is one-sided. Joint adjustment gives \(q_{\mathrm{BH}}=0.0234\) and \(p_{\mathrm{Holm}}=0.0502\) for every comparison. Contraction is the median first-to-last difference; parentheses count declines across datasets. Units are macro-F1 points (classification) and relative RMSE points (regression).}
		\label{tab:ordered-tests}
	\end{table}

	\subsection{The remaining gains are sparse.}
	Across dataset effects vary substantially by intervention. Numerical transformations are generally concentrated near zero, whereas binning, supervised selection, and most TabPrep variants are predominantly negative. Few interventions combine a positive across dataset median with improvements on at least half of their applicable datasets. Positive gains therefore reflect particular dataset–intervention–estimator combinations rather than methods that transfer consistently across tasks.
	
	\subsection{Transfer benefit depends on source--target comparability.}
	For each completed dataset--estimator cell, we compare target-only inference with inference augmented by NNRC-selected source samples under two target representations: the best-observed feature-engineering condition and the identity condition without external feature engineering. The paired difference therefore estimates the incremental value of selected related observations within a fixed representation, not the value of feature engineering itself. Each gain is the arithmetic mean of 30 paired split-level gains (10 repeats \(\times\) 3 folds). The best-observed feature condition remains an oracle descriptive choice, and the NNRC settings were not selected by an independent validation procedure. Airfoil benefits under both representations for TabPFN, whereas TabICL benefits only with numeric interactions. Concrete and maternal effects are smaller and representation dependent, consistent with their residual compatibility (Table~\ref{tab:source-pairs}); complete NNRC values are retained in the supplementary artifact.
	
	\subsection{No universally effective transformations.}
	For the latest evaluated models, all reported family-level medians are non-positive. Binning has median effects of -0.88 and -0.96 macro-F1 points for TabPFN v3 and TabICL v2 classification, and -6.70\% and -7.40\% relative RMSE reduction for regression. TabPrep is also negative in all four settings, with median effects ranging from -1.47 macro-F1 points to -4.13\% relative RMSE reduction. Numerical transformations are less disruptive, with medians close to zero, but their positive-cell frequencies vary from 1/35 in TabICL classification to 18/25 in TabICL regression. Thus, near-zero aggregate effects do not imply invariance across datasets. Overall, generic feature engineering offers less aggregate headroom in later model versions, but it is neither uniformly ineffective nor uniformly harmful.
	
	\section{Discussion}
	Our results indicate that the role of intervention is changing as tabular foundation models become stronger. Across the examined generations, the best observed gain from feature engineering declines. The typical intervention remains non positive. For the latest models, no evaluated feature family produces a consistently positive aggregate effect. The remaining gains occur only in specific combinations of datasets, transformations, and estimators. Results from earlier model generations therefore do not fully predict the behavior of later models. The small data regime is deliberate: it matches the intended use case of TFMs and provides a stringent test of whether feature engineering supplies useful inductive bias when task-specific evidence is limited. The observed decline in feature-engineering headroom therefore occurs in a setting where such interventions have substantial opportunity to help. We nevertheless scope our conclusions to the evaluated small-data benchmarks, with datasets—not repeated splits—as the independent units of inference
	
	A plausible explanation is that stronger TFMs extract more predictive structure directly from the original table. This reduces the value of generic transformations. The negative effects of binning, feature selection, and several composite pipelines show that later TFMs still respond to changes in representation. However, these changes may remove useful variation or create inputs that are less compatible with pretrained inference. Representation still matters, but generic re representation offers less benefit. Task specific transformations may remain useful when they provide structure that the model cannot recover from the original inputs.
	
	The context augmentation results provide a complementary view. Feature engineering reorganizes information already present in the target data. Context augmentation can introduce new task relevant information. Related source samples are one way to provide such information. Their benefit is clearest for airfoil. The effects are smaller or depend on the model for concrete and maternal health. This variation shows that the value of additional context depends strongly on source target compatibility. The key question is therefore not only how to transform the observed table, but also what additional information should be provided and how it should be organized.
	
	Taken together, our results point to a shift in the main source of improvement for stronger TFMs. Across the evaluated generations, generic transformations of existing inputs provide progressively less aggregate headroom. Compatible task relevant context can still produce measurable gains. The improvement frontier may therefore be moving from representation engineering toward information augmentation.
	
	\bibliography{references}

\onecolumn
\appendix
\section*{Exhaustive Raw-Metric Tables}
\label{sec:raw-metrics}
\noindent This appendix reports the full per-dataset raw-metric tables for every estimator and feature-engineering condition evaluated in the paper. Each table lists the condition mean and sample standard deviation across the 30 paired splits (10 repeats $\times$ 3 folds). Cells marked ``---'' indicate that the condition is not applicable to that dataset.
\begingroup
\scriptsize
\setlength{\tabcolsep}{1.35mm}
\begin{longtable}{@{}lccccc@{}}
\caption{Complete raw metric results for Airfoil noise, metric r2. Values are mean $\mathbin{\pm}$ sample standard deviation across the 30 paired splits.}\label{tab:supp-raw-1} \\
\toprule
Condition & PFN v2 & PFN v2.5 & PFN v2.6 & PFN v3 & ICL v2 \\
\midrule
\endfirsthead
\toprule
Condition & PFN v2 & PFN v2.5 & PFN v2.6 & PFN v3 & ICL v2 \\
\midrule
\endhead
\midrule
\multicolumn{6}{r}{\scriptsize Continued on next page} \\
\endfoot
\bottomrule
\endlastfoot
Identity & $0.9737 \mathbin{\pm} 0.0046$ & $0.9726 \mathbin{\pm} 0.0048$ & $0.9751 \mathbin{\pm} 0.0048$ & $0.9784 \mathbin{\pm} 0.0036$ & $0.9756 \mathbin{\pm} 0.0039$ \\
\addlinespace
Standard scaling & $0.9731 \mathbin{\pm} 0.0050$ & $0.9728 \mathbin{\pm} 0.0050$ & $0.9731 \mathbin{\pm} 0.0042$ & $0.9782 \mathbin{\pm} 0.0035$ & $0.9756 \mathbin{\pm} 0.0039$ \\
Log transform & $0.9730 \mathbin{\pm} 0.0046$ & $0.9725 \mathbin{\pm} 0.0051$ & $0.9722 \mathbin{\pm} 0.0046$ & $0.9781 \mathbin{\pm} 0.0038$ & $0.9763 \mathbin{\pm} 0.0040$ \\
Square-root transform & $0.9730 \mathbin{\pm} 0.0047$ & $0.9727 \mathbin{\pm} 0.0050$ & $0.9752 \mathbin{\pm} 0.0046$ & $0.9778 \mathbin{\pm} 0.0039$ & $0.9760 \mathbin{\pm} 0.0040$ \\
Box–Cox & $0.9730 \mathbin{\pm} 0.0047$ & $0.9726 \mathbin{\pm} 0.0050$ & $0.9721 \mathbin{\pm} 0.0048$ & $0.9785 \mathbin{\pm} 0.0034$ & $0.9759 \mathbin{\pm} 0.0038$ \\
Min–max scaling & $0.9733 \mathbin{\pm} 0.0047$ & $0.9728 \mathbin{\pm} 0.0050$ & $0.9733 \mathbin{\pm} 0.0049$ & $0.9782 \mathbin{\pm} 0.0036$ & $0.9756 \mathbin{\pm} 0.0039$ \\
\addlinespace
One-hot encoding & $0.9728 \mathbin{\pm} 0.0042$ & $0.9726 \mathbin{\pm} 0.0047$ & $0.9724 \mathbin{\pm} 0.0043$ & $0.9778 \mathbin{\pm} 0.0036$ & $0.9731 \mathbin{\pm} 0.0042$ \\
Ordinal encoding & $0.9731 \mathbin{\pm} 0.0049$ & $0.9728 \mathbin{\pm} 0.0044$ & $0.9750 \mathbin{\pm} 0.0046$ & $0.9782 \mathbin{\pm} 0.0036$ & $0.9754 \mathbin{\pm} 0.0041$ \\
\end{longtable}
\endgroup

\begingroup
\scriptsize
\setlength{\tabcolsep}{1.35mm}
\begin{longtable}{@{}lccccc@{}}
\caption{Complete raw metric results for Airfoil noise, metric rmse. Values are mean $\mathbin{\pm}$ sample standard deviation across the 30 paired splits.}\label{tab:supp-raw-2} \\
\toprule
Condition & PFN v2 & PFN v2.5 & PFN v2.6 & PFN v3 & ICL v2 \\
\midrule
\endfirsthead
\toprule
Condition & PFN v2 & PFN v2.5 & PFN v2.6 & PFN v3 & ICL v2 \\
\midrule
\endhead
\midrule
\multicolumn{6}{r}{\scriptsize Continued on next page} \\
\endfoot
\bottomrule
\endlastfoot
Identity & $1.1122 \mathbin{\pm} 0.0925$ & $1.1342 \mathbin{\pm} 0.0971$ & $1.0820 \mathbin{\pm} 0.1002$ & $1.0082 \mathbin{\pm} 0.0787$ & $1.0729 \mathbin{\pm} 0.0787$ \\
\addlinespace
Standard scaling & $1.1233 \mathbin{\pm} 0.0963$ & $1.1310 \mathbin{\pm} 0.0980$ & $1.1256 \mathbin{\pm} 0.0819$ & $1.0120 \mathbin{\pm} 0.0759$ & $1.0729 \mathbin{\pm} 0.0787$ \\
Log transform & $1.1277 \mathbin{\pm} 0.0889$ & $1.1365 \mathbin{\pm} 0.1010$ & $1.1448 \mathbin{\pm} 0.0938$ & $1.0157 \mathbin{\pm} 0.0830$ & $1.0572 \mathbin{\pm} 0.0832$ \\
Square-root transform & $1.1269 \mathbin{\pm} 0.0933$ & $1.1334 \mathbin{\pm} 0.0981$ & $1.0797 \mathbin{\pm} 0.0978$ & $1.0218 \mathbin{\pm} 0.0837$ & $1.0633 \mathbin{\pm} 0.0836$ \\
Box–Cox & $1.1276 \mathbin{\pm} 0.0899$ & $1.1342 \mathbin{\pm} 0.1003$ & $1.1456 \mathbin{\pm} 0.0936$ & $1.0062 \mathbin{\pm} 0.0750$ & $1.0664 \mathbin{\pm} 0.0774$ \\
Min–max scaling & $1.1213 \mathbin{\pm} 0.0916$ & $1.1303 \mathbin{\pm} 0.0980$ & $1.1208 \mathbin{\pm} 0.1051$ & $1.0128 \mathbin{\pm} 0.0794$ & $1.0729 \mathbin{\pm} 0.0787$ \\
\addlinespace
One-hot encoding & $1.1324 \mathbin{\pm} 0.0838$ & $1.1349 \mathbin{\pm} 0.0924$ & $1.1393 \mathbin{\pm} 0.0832$ & $1.0214 \mathbin{\pm} 0.0800$ & $1.1265 \mathbin{\pm} 0.0826$ \\
Ordinal encoding & $1.1237 \mathbin{\pm} 0.0949$ & $1.1307 \mathbin{\pm} 0.0871$ & $1.0833 \mathbin{\pm} 0.0989$ & $1.0139 \mathbin{\pm} 0.0791$ & $1.0751 \mathbin{\pm} 0.0848$ \\
\end{longtable}
\endgroup

\begingroup
\scriptsize
\setlength{\tabcolsep}{0.65mm}
\begin{longtable}{@{}lccccccc@{}}
\caption{Complete TabPFN and TabICL raw metric results for Anneal, metric accuracy. Values are mean $\mathbin{\pm}$ sample standard deviation across the 30 paired splits.}\label{tab:supp-raw-3}\label{tab:supp-raw-4} \\
\toprule
Condition & PFN v1 & PFN v2 & PFN v2.5 & PFN v2.6 & PFN v3 & ICL v1 & ICL v2 \\
\midrule
\endfirsthead
\toprule
Condition & PFN v1 & PFN v2 & PFN v2.5 & PFN v2.6 & PFN v3 & ICL v1 & ICL v2 \\
\midrule
\endhead
\midrule
\multicolumn{8}{r}{\scriptsize Continued on next page} \\
\endfoot
\bottomrule
\endlastfoot
Identity & $0.9709 \mathbin{\pm} 0.0106$ & $0.9970 \mathbin{\pm} 0.0045$ & $0.9969 \mathbin{\pm} 0.0044$ & $0.9951 \mathbin{\pm} 0.0062$ & $0.9968 \mathbin{\pm} 0.0042$ & $0.9885 \mathbin{\pm} 0.0065$ & $0.9952 \mathbin{\pm} 0.0052$ \\
\addlinespace
Standard scaling & $0.9703 \mathbin{\pm} 0.0099$ & $0.9968 \mathbin{\pm} 0.0046$ & $0.9967 \mathbin{\pm} 0.0044$ & $0.9952 \mathbin{\pm} 0.0062$ & $0.9968 \mathbin{\pm} 0.0038$ & $0.9882 \mathbin{\pm} 0.0067$ & $0.9952 \mathbin{\pm} 0.0052$ \\
Log transform & $0.9619 \mathbin{\pm} 0.0123$ & $0.9964 \mathbin{\pm} 0.0055$ & $0.9970 \mathbin{\pm} 0.0038$ & $0.9951 \mathbin{\pm} 0.0064$ & $0.9941 \mathbin{\pm} 0.0067$ & $0.9806 \mathbin{\pm} 0.0076$ & $0.9952 \mathbin{\pm} 0.0056$ \\
Square-root transform & $0.9651 \mathbin{\pm} 0.0115$ & $0.9969 \mathbin{\pm} 0.0046$ & $0.9970 \mathbin{\pm} 0.0040$ & $0.9950 \mathbin{\pm} 0.0064$ & $0.9963 \mathbin{\pm} 0.0048$ & $0.9866 \mathbin{\pm} 0.0060$ & $0.9948 \mathbin{\pm} 0.0056$ \\
Box–Cox & $0.9710 \mathbin{\pm} 0.0105$ & $0.9967 \mathbin{\pm} 0.0045$ & $0.9967 \mathbin{\pm} 0.0044$ & $0.9953 \mathbin{\pm} 0.0067$ & $0.9965 \mathbin{\pm} 0.0046$ & $0.9883 \mathbin{\pm} 0.0059$ & $0.9952 \mathbin{\pm} 0.0050$ \\
Min–max scaling & $0.9703 \mathbin{\pm} 0.0099$ & $0.9969 \mathbin{\pm} 0.0046$ & $0.9967 \mathbin{\pm} 0.0044$ & $0.9953 \mathbin{\pm} 0.0062$ & $0.9969 \mathbin{\pm} 0.0037$ & $0.9882 \mathbin{\pm} 0.0067$ & $0.9952 \mathbin{\pm} 0.0052$ \\
\addlinespace
One-hot encoding & $0.8939 \mathbin{\pm} 0.0115$ & $0.9156 \mathbin{\pm} 0.0099$ & $0.9174 \mathbin{\pm} 0.0099$ & $0.9177 \mathbin{\pm} 0.0086$ & $0.9130 \mathbin{\pm} 0.0116$ & $0.9000 \mathbin{\pm} 0.0101$ & $0.9223 \mathbin{\pm} 0.0089$ \\
Ordinal encoding & $0.9706 \mathbin{\pm} 0.0111$ & $0.9968 \mathbin{\pm} 0.0045$ & $0.9970 \mathbin{\pm} 0.0045$ & $0.9953 \mathbin{\pm} 0.0061$ & $0.9967 \mathbin{\pm} 0.0047$ & $0.9876 \mathbin{\pm} 0.0066$ & $0.9952 \mathbin{\pm} 0.0055$ \\
\end{longtable}
\endgroup

\begingroup
\scriptsize
\setlength{\tabcolsep}{0.65mm}
\begin{longtable}{@{}lccccccc@{}}
\caption{Complete TabPFN and TabICL raw metric results for Anneal, metric macro f1. Values are mean $\mathbin{\pm}$ sample standard deviation across the 30 paired splits.}\label{tab:supp-raw-5}\label{tab:supp-raw-6} \\
\toprule
Condition & PFN v1 & PFN v2 & PFN v2.5 & PFN v2.6 & PFN v3 & ICL v1 & ICL v2 \\
\midrule
\endfirsthead
\toprule
Condition & PFN v1 & PFN v2 & PFN v2.5 & PFN v2.6 & PFN v3 & ICL v1 & ICL v2 \\
\midrule
\endhead
\midrule
\multicolumn{8}{r}{\scriptsize Continued on next page} \\
\endfoot
\bottomrule
\endlastfoot
Identity & $0.8755 \mathbin{\pm} 0.0688$ & $0.9715 \mathbin{\pm} 0.0466$ & $0.9714 \mathbin{\pm} 0.0465$ & $0.9712 \mathbin{\pm} 0.0491$ & $0.9771 \mathbin{\pm} 0.0426$ & $0.9542 \mathbin{\pm} 0.0459$ & $0.9721 \mathbin{\pm} 0.0473$ \\
\addlinespace
Standard scaling & $0.8663 \mathbin{\pm} 0.0726$ & $0.9711 \mathbin{\pm} 0.0467$ & $0.9672 \mathbin{\pm} 0.0484$ & $0.9713 \mathbin{\pm} 0.0491$ & $0.9773 \mathbin{\pm} 0.0412$ & $0.9505 \mathbin{\pm} 0.0457$ & $0.9721 \mathbin{\pm} 0.0473$ \\
Log transform & $0.8682 \mathbin{\pm} 0.0641$ & $0.9721 \mathbin{\pm} 0.0471$ & $0.9731 \mathbin{\pm} 0.0428$ & $0.9712 \mathbin{\pm} 0.0492$ & $0.9713 \mathbin{\pm} 0.0457$ & $0.9327 \mathbin{\pm} 0.0470$ & $0.9720 \mathbin{\pm} 0.0478$ \\
Square-root transform & $0.8603 \mathbin{\pm} 0.0745$ & $0.9714 \mathbin{\pm} 0.0466$ & $0.9717 \mathbin{\pm} 0.0450$ & $0.9711 \mathbin{\pm} 0.0492$ & $0.9739 \mathbin{\pm} 0.0438$ & $0.9470 \mathbin{\pm} 0.0434$ & $0.9717 \mathbin{\pm} 0.0478$ \\
Box–Cox & $0.8683 \mathbin{\pm} 0.0754$ & $0.9699 \mathbin{\pm} 0.0463$ & $0.9672 \mathbin{\pm} 0.0484$ & $0.9709 \mathbin{\pm} 0.0506$ & $0.9737 \mathbin{\pm} 0.0459$ & $0.9475 \mathbin{\pm} 0.0534$ & $0.9721 \mathbin{\pm} 0.0472$ \\
Min–max scaling & $0.8663 \mathbin{\pm} 0.0726$ & $0.9714 \mathbin{\pm} 0.0466$ & $0.9672 \mathbin{\pm} 0.0484$ & $0.9714 \mathbin{\pm} 0.0491$ & $0.9774 \mathbin{\pm} 0.0412$ & $0.9505 \mathbin{\pm} 0.0457$ & $0.9721 \mathbin{\pm} 0.0473$ \\
\addlinespace
One-hot encoding & $0.6766 \mathbin{\pm} 0.0498$ & $0.7547 \mathbin{\pm} 0.0589$ & $0.7596 \mathbin{\pm} 0.0582$ & $0.7570 \mathbin{\pm} 0.0594$ & $0.7484 \mathbin{\pm} 0.0601$ & $0.7221 \mathbin{\pm} 0.0573$ & $0.7646 \mathbin{\pm} 0.0576$ \\
Ordinal encoding & $0.8825 \mathbin{\pm} 0.0733$ & $0.9711 \mathbin{\pm} 0.0465$ & $0.9713 \mathbin{\pm} 0.0465$ & $0.9714 \mathbin{\pm} 0.0488$ & $0.9739 \mathbin{\pm} 0.0465$ & $0.9466 \mathbin{\pm} 0.0459$ & $0.9721 \mathbin{\pm} 0.0475$ \\
\end{longtable}
\endgroup

\begingroup
\scriptsize
\setlength{\tabcolsep}{1.35mm}
\begin{longtable}{@{}lccccc@{}}
\caption{Complete raw metric results for Used Fiat 500, metric r2. Values are mean $\mathbin{\pm}$ sample standard deviation across the 30 paired splits.}\label{tab:supp-raw-7} \\
\toprule
Condition & PFN v2 & PFN v2.5 & PFN v2.6 & PFN v3 & ICL v2 \\
\midrule
\endfirsthead
\toprule
Condition & PFN v2 & PFN v2.5 & PFN v2.6 & PFN v3 & ICL v2 \\
\midrule
\endhead
\midrule
\multicolumn{6}{r}{\scriptsize Continued on next page} \\
\endfoot
\bottomrule
\endlastfoot
Identity & $0.8587 \mathbin{\pm} 0.0099$ & $0.8578 \mathbin{\pm} 0.0105$ & $0.8589 \mathbin{\pm} 0.0110$ & $0.8637 \mathbin{\pm} 0.0104$ & $0.8632 \mathbin{\pm} 0.0102$ \\
\addlinespace
Standard scaling & $0.8587 \mathbin{\pm} 0.0101$ & $0.8578 \mathbin{\pm} 0.0104$ & $0.8601 \mathbin{\pm} 0.0109$ & $0.8637 \mathbin{\pm} 0.0103$ & $0.8632 \mathbin{\pm} 0.0102$ \\
Log transform & $0.8583 \mathbin{\pm} 0.0096$ & $0.8577 \mathbin{\pm} 0.0103$ & $0.8587 \mathbin{\pm} 0.0110$ & $0.8630 \mathbin{\pm} 0.0104$ & $0.8634 \mathbin{\pm} 0.0101$ \\
Square-root transform & $0.8584 \mathbin{\pm} 0.0099$ & $0.8576 \mathbin{\pm} 0.0104$ & $0.8589 \mathbin{\pm} 0.0111$ & $0.8631 \mathbin{\pm} 0.0105$ & $0.8634 \mathbin{\pm} 0.0101$ \\
Box–Cox & $0.8581 \mathbin{\pm} 0.0100$ & $0.8571 \mathbin{\pm} 0.0104$ & $0.8600 \mathbin{\pm} 0.0107$ & $0.8630 \mathbin{\pm} 0.0104$ & $0.8632 \mathbin{\pm} 0.0102$ \\
Min–max scaling & $0.8592 \mathbin{\pm} 0.0101$ & $0.8577 \mathbin{\pm} 0.0105$ & $0.8608 \mathbin{\pm} 0.0111$ & $0.8637 \mathbin{\pm} 0.0103$ & $0.8632 \mathbin{\pm} 0.0102$ \\
\addlinespace
One-hot encoding & $0.8587 \mathbin{\pm} 0.0099$ & $0.8585 \mathbin{\pm} 0.0101$ & $0.8599 \mathbin{\pm} 0.0107$ & $0.8637 \mathbin{\pm} 0.0103$ & $0.8633 \mathbin{\pm} 0.0098$ \\
Ordinal encoding & $0.8587 \mathbin{\pm} 0.0099$ & $0.8576 \mathbin{\pm} 0.0104$ & $0.8589 \mathbin{\pm} 0.0110$ & $0.8638 \mathbin{\pm} 0.0102$ & $0.8632 \mathbin{\pm} 0.0102$ \\
\end{longtable}
\endgroup

\begingroup
\scriptsize
\setlength{\tabcolsep}{1.35mm}
\begin{longtable}{@{}lccccc@{}}
\caption{Complete raw metric results for Used Fiat 500, metric rmse. Values are mean $\mathbin{\pm}$ sample standard deviation across the 30 paired splits.}\label{tab:supp-raw-8} \\
\toprule
Condition & PFN v2 & PFN v2.5 & PFN v2.6 & PFN v3 & ICL v2 \\
\midrule
\endfirsthead
\toprule
Condition & PFN v2 & PFN v2.5 & PFN v2.6 & PFN v3 & ICL v2 \\
\midrule
\endhead
\midrule
\multicolumn{6}{r}{\scriptsize Continued on next page} \\
\endfoot
\bottomrule
\endlastfoot
Identity & $727.4 \mathbin{\pm} 23.4$ & $729.7 \mathbin{\pm} 23.8$ & $726.8 \mathbin{\pm} 24.8$ & $714.4 \mathbin{\pm} 23.2$ & $715.7 \mathbin{\pm} 23.4$ \\
\addlinespace
Standard scaling & $727.6 \mathbin{\pm} 24.0$ & $729.9 \mathbin{\pm} 23.8$ & $723.7 \mathbin{\pm} 24.9$ & $714.5 \mathbin{\pm} 23.2$ & $715.7 \mathbin{\pm} 23.4$ \\
Log transform & $728.6 \mathbin{\pm} 23.0$ & $730.1 \mathbin{\pm} 23.5$ & $727.5 \mathbin{\pm} 24.6$ & $716.4 \mathbin{\pm} 23.4$ & $715.4 \mathbin{\pm} 23.2$ \\
Square-root transform & $728.2 \mathbin{\pm} 23.4$ & $730.3 \mathbin{\pm} 23.6$ & $726.8 \mathbin{\pm} 25.0$ & $716.1 \mathbin{\pm} 23.4$ & $715.2 \mathbin{\pm} 23.1$ \\
Box–Cox & $729.1 \mathbin{\pm} 23.8$ & $731.7 \mathbin{\pm} 23.5$ & $724.0 \mathbin{\pm} 24.5$ & $716.4 \mathbin{\pm} 23.8$ & $715.7 \mathbin{\pm} 23.6$ \\
Min–max scaling & $726.3 \mathbin{\pm} 24.1$ & $729.9 \mathbin{\pm} 23.8$ & $722.0 \mathbin{\pm} 25.0$ & $714.3 \mathbin{\pm} 23.1$ & $715.7 \mathbin{\pm} 23.4$ \\
\addlinespace
One-hot encoding & $727.6 \mathbin{\pm} 23.5$ & $728.0 \mathbin{\pm} 23.2$ & $724.2 \mathbin{\pm} 24.2$ & $714.5 \mathbin{\pm} 23.0$ & $715.4 \mathbin{\pm} 22.8$ \\
Ordinal encoding & $727.4 \mathbin{\pm} 23.4$ & $730.2 \mathbin{\pm} 23.2$ & $726.9 \mathbin{\pm} 25.0$ & $714.2 \mathbin{\pm} 23.0$ & $715.7 \mathbin{\pm} 23.2$ \\
\end{longtable}
\endgroup

\begingroup
\scriptsize
\setlength{\tabcolsep}{0.65mm}
\begin{longtable}{@{}lccccccc@{}}
\caption{Complete TabPFN and TabICL raw metric results for Blood transfusion, metric accuracy. Values are mean $\mathbin{\pm}$ sample standard deviation across the 30 paired splits.}\label{tab:supp-raw-9}\label{tab:supp-raw-10} \\
\toprule
Condition & PFN v1 & PFN v2 & PFN v2.5 & PFN v2.6 & PFN v3 & ICL v1 & ICL v2 \\
\midrule
\endfirsthead
\toprule
Condition & PFN v1 & PFN v2 & PFN v2.5 & PFN v2.6 & PFN v3 & ICL v1 & ICL v2 \\
\midrule
\endhead
\midrule
\multicolumn{8}{r}{\scriptsize Continued on next page} \\
\endfoot
\bottomrule
\endlastfoot
Identity & $0.7925 \mathbin{\pm} 0.0113$ & $0.7898 \mathbin{\pm} 0.0099$ & $0.7932 \mathbin{\pm} 0.0090$ & $0.7894 \mathbin{\pm} 0.0111$ & $0.7904 \mathbin{\pm} 0.0102$ & $0.7918 \mathbin{\pm} 0.0086$ & $0.7913 \mathbin{\pm} 0.0107$ \\
\addlinespace
Standard scaling & $0.7925 \mathbin{\pm} 0.0113$ & $0.7900 \mathbin{\pm} 0.0093$ & $0.7928 \mathbin{\pm} 0.0096$ & $0.7894 \mathbin{\pm} 0.0111$ & $0.7901 \mathbin{\pm} 0.0096$ & $0.7918 \mathbin{\pm} 0.0086$ & $0.7913 \mathbin{\pm} 0.0107$ \\
Log transform & $0.7901 \mathbin{\pm} 0.0117$ & $0.7878 \mathbin{\pm} 0.0100$ & $0.7885 \mathbin{\pm} 0.0105$ & $0.7892 \mathbin{\pm} 0.0096$ & $0.7877 \mathbin{\pm} 0.0127$ & $0.7874 \mathbin{\pm} 0.0105$ & $0.7913 \mathbin{\pm} 0.0105$ \\
Square-root transform & $0.7911 \mathbin{\pm} 0.0115$ & $0.7886 \mathbin{\pm} 0.0103$ & $0.7901 \mathbin{\pm} 0.0106$ & $0.7878 \mathbin{\pm} 0.0121$ & $0.7885 \mathbin{\pm} 0.0114$ & $0.7889 \mathbin{\pm} 0.0081$ & $0.7911 \mathbin{\pm} 0.0099$ \\
Box–Cox & $0.7928 \mathbin{\pm} 0.0122$ & $0.7870 \mathbin{\pm} 0.0107$ & $0.7904 \mathbin{\pm} 0.0098$ & $0.7893 \mathbin{\pm} 0.0112$ & $0.7882 \mathbin{\pm} 0.0105$ & $0.7888 \mathbin{\pm} 0.0098$ & $0.7919 \mathbin{\pm} 0.0104$ \\
Min–max scaling & $0.7925 \mathbin{\pm} 0.0113$ & $0.7890 \mathbin{\pm} 0.0090$ & $0.7933 \mathbin{\pm} 0.0107$ & $0.7894 \mathbin{\pm} 0.0111$ & $0.7897 \mathbin{\pm} 0.0113$ & $0.7920 \mathbin{\pm} 0.0086$ & $0.7913 \mathbin{\pm} 0.0107$ \\
\addlinespace
One-hot encoding & --- & --- & --- & --- & --- & --- & --- \\
Ordinal encoding & --- & --- & --- & --- & --- & --- & --- \\
\end{longtable}
\endgroup

\begingroup
\scriptsize
\setlength{\tabcolsep}{0.65mm}
\begin{longtable}{@{}lccccccc@{}}
\caption{Complete TabPFN and TabICL raw metric results for Blood transfusion, metric macro f1. Values are mean $\mathbin{\pm}$ sample standard deviation across the 30 paired splits.}\label{tab:supp-raw-11}\label{tab:supp-raw-12} \\
\toprule
Condition & PFN v1 & PFN v2 & PFN v2.5 & PFN v2.6 & PFN v3 & ICL v1 & ICL v2 \\
\midrule
\endfirsthead
\toprule
Condition & PFN v1 & PFN v2 & PFN v2.5 & PFN v2.6 & PFN v3 & ICL v1 & ICL v2 \\
\midrule
\endhead
\midrule
\multicolumn{8}{r}{\scriptsize Continued on next page} \\
\endfoot
\bottomrule
\endlastfoot
Identity & $0.6452 \mathbin{\pm} 0.0321$ & $0.6232 \mathbin{\pm} 0.0362$ & $0.6367 \mathbin{\pm} 0.0382$ & $0.6359 \mathbin{\pm} 0.0335$ & $0.6290 \mathbin{\pm} 0.0398$ & $0.6368 \mathbin{\pm} 0.0361$ & $0.6389 \mathbin{\pm} 0.0357$ \\
\addlinespace
Standard scaling & $0.6452 \mathbin{\pm} 0.0321$ & $0.6217 \mathbin{\pm} 0.0370$ & $0.6360 \mathbin{\pm} 0.0370$ & $0.6359 \mathbin{\pm} 0.0335$ & $0.6273 \mathbin{\pm} 0.0404$ & $0.6368 \mathbin{\pm} 0.0361$ & $0.6389 \mathbin{\pm} 0.0357$ \\
Log transform & $0.6378 \mathbin{\pm} 0.0335$ & $0.6208 \mathbin{\pm} 0.0386$ & $0.6290 \mathbin{\pm} 0.0347$ & $0.6359 \mathbin{\pm} 0.0323$ & $0.6255 \mathbin{\pm} 0.0394$ & $0.6226 \mathbin{\pm} 0.0415$ & $0.6374 \mathbin{\pm} 0.0361$ \\
Square-root transform & $0.6402 \mathbin{\pm} 0.0326$ & $0.6189 \mathbin{\pm} 0.0347$ & $0.6315 \mathbin{\pm} 0.0304$ & $0.6345 \mathbin{\pm} 0.0333$ & $0.6271 \mathbin{\pm} 0.0401$ & $0.6286 \mathbin{\pm} 0.0396$ & $0.6378 \mathbin{\pm} 0.0342$ \\
Box–Cox & $0.6465 \mathbin{\pm} 0.0329$ & $0.6155 \mathbin{\pm} 0.0424$ & $0.6321 \mathbin{\pm} 0.0344$ & $0.6376 \mathbin{\pm} 0.0374$ & $0.6237 \mathbin{\pm} 0.0398$ & $0.6260 \mathbin{\pm} 0.0431$ & $0.6397 \mathbin{\pm} 0.0353$ \\
Min–max scaling & $0.6452 \mathbin{\pm} 0.0321$ & $0.6203 \mathbin{\pm} 0.0395$ & $0.6363 \mathbin{\pm} 0.0394$ & $0.6359 \mathbin{\pm} 0.0335$ & $0.6270 \mathbin{\pm} 0.0411$ & $0.6371 \mathbin{\pm} 0.0358$ & $0.6389 \mathbin{\pm} 0.0357$ \\
\addlinespace
One-hot encoding & --- & --- & --- & --- & --- & --- & --- \\
Ordinal encoding & --- & --- & --- & --- & --- & --- & --- \\
\end{longtable}
\endgroup

\begingroup
\scriptsize
\setlength{\tabcolsep}{1.35mm}
\begin{longtable}{@{}lccccc@{}}
\caption{Complete raw metric results for Concrete strength, metric r2. Values are mean $\mathbin{\pm}$ sample standard deviation across the 30 paired splits.}\label{tab:supp-raw-13} \\
\toprule
Condition & PFN v2 & PFN v2.5 & PFN v2.6 & PFN v3 & ICL v2 \\
\midrule
\endfirsthead
\toprule
Condition & PFN v2 & PFN v2.5 & PFN v2.6 & PFN v3 & ICL v2 \\
\midrule
\endhead
\midrule
\multicolumn{6}{r}{\scriptsize Continued on next page} \\
\endfoot
\bottomrule
\endlastfoot
Identity & $0.9347 \mathbin{\pm} 0.0126$ & $0.9394 \mathbin{\pm} 0.0124$ & $0.9405 \mathbin{\pm} 0.0119$ & $0.9411 \mathbin{\pm} 0.0125$ & $0.9424 \mathbin{\pm} 0.0133$ \\
\addlinespace
Standard scaling & $0.9338 \mathbin{\pm} 0.0128$ & $0.9393 \mathbin{\pm} 0.0124$ & $0.9387 \mathbin{\pm} 0.0118$ & $0.9411 \mathbin{\pm} 0.0125$ & $0.9424 \mathbin{\pm} 0.0133$ \\
Log transform & $0.9344 \mathbin{\pm} 0.0130$ & $0.9400 \mathbin{\pm} 0.0124$ & $0.9399 \mathbin{\pm} 0.0121$ & $0.9427 \mathbin{\pm} 0.0124$ & $0.9430 \mathbin{\pm} 0.0128$ \\
Square-root transform & $0.9347 \mathbin{\pm} 0.0127$ & $0.9402 \mathbin{\pm} 0.0127$ & $0.9402 \mathbin{\pm} 0.0120$ & $0.9421 \mathbin{\pm} 0.0127$ & $0.9425 \mathbin{\pm} 0.0132$ \\
Box–Cox & $0.9352 \mathbin{\pm} 0.0125$ & $0.9413 \mathbin{\pm} 0.0121$ & $0.9409 \mathbin{\pm} 0.0124$ & $0.9419 \mathbin{\pm} 0.0123$ & $0.9433 \mathbin{\pm} 0.0129$ \\
Min–max scaling & $0.9339 \mathbin{\pm} 0.0129$ & $0.9394 \mathbin{\pm} 0.0125$ & $0.9403 \mathbin{\pm} 0.0121$ & $0.9412 \mathbin{\pm} 0.0125$ & $0.9424 \mathbin{\pm} 0.0133$ \\
\addlinespace
One-hot encoding & --- & --- & --- & --- & --- \\
Ordinal encoding & --- & --- & --- & --- & --- \\
\end{longtable}
\endgroup

\begingroup
\scriptsize
\setlength{\tabcolsep}{1.35mm}
\begin{longtable}{@{}lccccc@{}}
\caption{Complete raw metric results for Concrete strength, metric rmse. Values are mean $\mathbin{\pm}$ sample standard deviation across the 30 paired splits.}\label{tab:supp-raw-14} \\
\toprule
Condition & PFN v2 & PFN v2.5 & PFN v2.6 & PFN v3 & ICL v2 \\
\midrule
\endfirsthead
\toprule
Condition & PFN v2 & PFN v2.5 & PFN v2.6 & PFN v3 & ICL v2 \\
\midrule
\endhead
\midrule
\multicolumn{6}{r}{\scriptsize Continued on next page} \\
\endfoot
\bottomrule
\endlastfoot
Identity & $4.2382 \mathbin{\pm} 0.3926$ & $4.0788 \mathbin{\pm} 0.3749$ & $4.0434 \mathbin{\pm} 0.3826$ & $4.0203 \mathbin{\pm} 0.4037$ & $3.9739 \mathbin{\pm} 0.4378$ \\
\addlinespace
Standard scaling & $4.2664 \mathbin{\pm} 0.3856$ & $4.0818 \mathbin{\pm} 0.3734$ & $4.1077 \mathbin{\pm} 0.3735$ & $4.0202 \mathbin{\pm} 0.4031$ & $3.9739 \mathbin{\pm} 0.4378$ \\
Log transform & $4.2486 \mathbin{\pm} 0.4000$ & $4.0588 \mathbin{\pm} 0.3778$ & $4.0637 \mathbin{\pm} 0.3854$ & $3.9656 \mathbin{\pm} 0.4067$ & $3.9543 \mathbin{\pm} 0.4253$ \\
Square-root transform & $4.2393 \mathbin{\pm} 0.3933$ & $4.0500 \mathbin{\pm} 0.3853$ & $4.0526 \mathbin{\pm} 0.3842$ & $3.9848 \mathbin{\pm} 0.4127$ & $3.9694 \mathbin{\pm} 0.4356$ \\
Box–Cox & $4.2222 \mathbin{\pm} 0.3829$ & $4.0160 \mathbin{\pm} 0.3751$ & $4.0297 \mathbin{\pm} 0.4026$ & $3.9920 \mathbin{\pm} 0.3979$ & $3.9420 \mathbin{\pm} 0.4290$ \\
Min–max scaling & $4.2651 \mathbin{\pm} 0.3934$ & $4.0799 \mathbin{\pm} 0.3765$ & $4.0495 \mathbin{\pm} 0.3826$ & $4.0193 \mathbin{\pm} 0.4025$ & $3.9739 \mathbin{\pm} 0.4378$ \\
\addlinespace
One-hot encoding & --- & --- & --- & --- & --- \\
Ordinal encoding & --- & --- & --- & --- & --- \\
\end{longtable}
\endgroup

\begingroup
\scriptsize
\setlength{\tabcolsep}{0.65mm}
\begin{longtable}{@{}lccccccc@{}}
\caption{Complete TabPFN and TabICL raw metric results for Credit-g, metric accuracy. Values are mean $\mathbin{\pm}$ sample standard deviation across the 30 paired splits.}\label{tab:supp-raw-15}\label{tab:supp-raw-16} \\
\toprule
Condition & PFN v1 & PFN v2 & PFN v2.5 & PFN v2.6 & PFN v3 & ICL v1 & ICL v2 \\
\midrule
\endfirsthead
\toprule
Condition & PFN v1 & PFN v2 & PFN v2.5 & PFN v2.6 & PFN v3 & ICL v1 & ICL v2 \\
\midrule
\endhead
\midrule
\multicolumn{8}{r}{\scriptsize Continued on next page} \\
\endfoot
\bottomrule
\endlastfoot
Identity & $0.7332 \mathbin{\pm} 0.0124$ & $0.7557 \mathbin{\pm} 0.0202$ & $0.7584 \mathbin{\pm} 0.0206$ & $0.7583 \mathbin{\pm} 0.0194$ & $0.7583 \mathbin{\pm} 0.0189$ & $0.7572 \mathbin{\pm} 0.0187$ & $0.7596 \mathbin{\pm} 0.0185$ \\
\addlinespace
Standard scaling & $0.7348 \mathbin{\pm} 0.0130$ & $0.7560 \mathbin{\pm} 0.0215$ & $0.7582 \mathbin{\pm} 0.0212$ & $0.7580 \mathbin{\pm} 0.0198$ & $0.7590 \mathbin{\pm} 0.0193$ & $0.7572 \mathbin{\pm} 0.0179$ & $0.7596 \mathbin{\pm} 0.0185$ \\
Log transform & $0.7342 \mathbin{\pm} 0.0138$ & $0.7565 \mathbin{\pm} 0.0238$ & $0.7582 \mathbin{\pm} 0.0197$ & $0.7597 \mathbin{\pm} 0.0203$ & $0.7604 \mathbin{\pm} 0.0181$ & $0.7554 \mathbin{\pm} 0.0165$ & $0.7584 \mathbin{\pm} 0.0211$ \\
Square-root transform & $0.7346 \mathbin{\pm} 0.0126$ & $0.7560 \mathbin{\pm} 0.0198$ & $0.7571 \mathbin{\pm} 0.0187$ & $0.7593 \mathbin{\pm} 0.0198$ & $0.7612 \mathbin{\pm} 0.0195$ & $0.7566 \mathbin{\pm} 0.0176$ & $0.7596 \mathbin{\pm} 0.0198$ \\
Box–Cox & $0.7331 \mathbin{\pm} 0.0147$ & $0.7547 \mathbin{\pm} 0.0257$ & $0.7582 \mathbin{\pm} 0.0205$ & $0.7588 \mathbin{\pm} 0.0210$ & $0.7591 \mathbin{\pm} 0.0197$ & $0.7557 \mathbin{\pm} 0.0172$ & $0.7575 \mathbin{\pm} 0.0198$ \\
Min–max scaling & $0.7348 \mathbin{\pm} 0.0130$ & $0.7561 \mathbin{\pm} 0.0213$ & $0.7582 \mathbin{\pm} 0.0206$ & $0.7579 \mathbin{\pm} 0.0198$ & $0.7607 \mathbin{\pm} 0.0194$ & $0.7569 \mathbin{\pm} 0.0183$ & $0.7596 \mathbin{\pm} 0.0185$ \\
\addlinespace
One-hot encoding & $0.7579 \mathbin{\pm} 0.0195$ & $0.7497 \mathbin{\pm} 0.0121$ & $0.7558 \mathbin{\pm} 0.0153$ & $0.7536 \mathbin{\pm} 0.0141$ & $0.7591 \mathbin{\pm} 0.0171$ & $0.7630 \mathbin{\pm} 0.0189$ & $0.7571 \mathbin{\pm} 0.0188$ \\
Ordinal encoding & $0.7341 \mathbin{\pm} 0.0134$ & $0.7535 \mathbin{\pm} 0.0174$ & $0.7579 \mathbin{\pm} 0.0180$ & $0.7585 \mathbin{\pm} 0.0174$ & $0.7599 \mathbin{\pm} 0.0189$ & $0.7581 \mathbin{\pm} 0.0155$ & $0.7595 \mathbin{\pm} 0.0187$ \\
\end{longtable}
\endgroup

\begingroup
\scriptsize
\setlength{\tabcolsep}{0.65mm}
\begin{longtable}{@{}lccccccc@{}}
\caption{Complete TabPFN and TabICL raw metric results for Credit-g, metric macro f1. Values are mean $\mathbin{\pm}$ sample standard deviation across the 30 paired splits.}\label{tab:supp-raw-17}\label{tab:supp-raw-18} \\
\toprule
Condition & PFN v1 & PFN v2 & PFN v2.5 & PFN v2.6 & PFN v3 & ICL v1 & ICL v2 \\
\midrule
\endfirsthead
\toprule
Condition & PFN v1 & PFN v2 & PFN v2.5 & PFN v2.6 & PFN v3 & ICL v1 & ICL v2 \\
\midrule
\endhead
\midrule
\multicolumn{8}{r}{\scriptsize Continued on next page} \\
\endfoot
\bottomrule
\endlastfoot
Identity & $0.6167 \mathbin{\pm} 0.0256$ & $0.6719 \mathbin{\pm} 0.0241$ & $0.6751 \mathbin{\pm} 0.0260$ & $0.6808 \mathbin{\pm} 0.0267$ & $0.6839 \mathbin{\pm} 0.0221$ & $0.6763 \mathbin{\pm} 0.0236$ & $0.6820 \mathbin{\pm} 0.0237$ \\
\addlinespace
Standard scaling & $0.6195 \mathbin{\pm} 0.0249$ & $0.6716 \mathbin{\pm} 0.0271$ & $0.6751 \mathbin{\pm} 0.0273$ & $0.6804 \mathbin{\pm} 0.0275$ & $0.6847 \mathbin{\pm} 0.0226$ & $0.6779 \mathbin{\pm} 0.0224$ & $0.6820 \mathbin{\pm} 0.0237$ \\
Log transform & $0.6204 \mathbin{\pm} 0.0230$ & $0.6746 \mathbin{\pm} 0.0295$ & $0.6763 \mathbin{\pm} 0.0256$ & $0.6827 \mathbin{\pm} 0.0275$ & $0.6865 \mathbin{\pm} 0.0216$ & $0.6745 \mathbin{\pm} 0.0213$ & $0.6805 \mathbin{\pm} 0.0263$ \\
Square-root transform & $0.6207 \mathbin{\pm} 0.0234$ & $0.6729 \mathbin{\pm} 0.0256$ & $0.6745 \mathbin{\pm} 0.0240$ & $0.6819 \mathbin{\pm} 0.0264$ & $0.6878 \mathbin{\pm} 0.0237$ & $0.6778 \mathbin{\pm} 0.0228$ & $0.6825 \mathbin{\pm} 0.0248$ \\
Box–Cox & $0.6187 \mathbin{\pm} 0.0248$ & $0.6727 \mathbin{\pm} 0.0318$ & $0.6753 \mathbin{\pm} 0.0264$ & $0.6816 \mathbin{\pm} 0.0288$ & $0.6855 \mathbin{\pm} 0.0228$ & $0.6734 \mathbin{\pm} 0.0225$ & $0.6803 \mathbin{\pm} 0.0243$ \\
Min–max scaling & $0.6195 \mathbin{\pm} 0.0249$ & $0.6718 \mathbin{\pm} 0.0263$ & $0.6748 \mathbin{\pm} 0.0276$ & $0.6801 \mathbin{\pm} 0.0277$ & $0.6871 \mathbin{\pm} 0.0225$ & $0.6775 \mathbin{\pm} 0.0230$ & $0.6820 \mathbin{\pm} 0.0237$ \\
\addlinespace
One-hot encoding & $0.6830 \mathbin{\pm} 0.0234$ & $0.6642 \mathbin{\pm} 0.0157$ & $0.6767 \mathbin{\pm} 0.0207$ & $0.6782 \mathbin{\pm} 0.0185$ & $0.6837 \mathbin{\pm} 0.0196$ & $0.6903 \mathbin{\pm} 0.0240$ & $0.6785 \mathbin{\pm} 0.0244$ \\
Ordinal encoding & $0.6173 \mathbin{\pm} 0.0265$ & $0.6677 \mathbin{\pm} 0.0228$ & $0.6749 \mathbin{\pm} 0.0234$ & $0.6811 \mathbin{\pm} 0.0227$ & $0.6862 \mathbin{\pm} 0.0229$ & $0.6797 \mathbin{\pm} 0.0206$ & $0.6818 \mathbin{\pm} 0.0224$ \\
\end{longtable}
\endgroup

\begingroup
\scriptsize
\setlength{\tabcolsep}{0.65mm}
\begin{longtable}{@{}lccccccc@{}}
\caption{Complete TabPFN and TabICL raw metric results for Diabetes, metric accuracy. Values are mean $\mathbin{\pm}$ sample standard deviation across the 30 paired splits.}\label{tab:supp-raw-19}\label{tab:supp-raw-20} \\
\toprule
Condition & PFN v1 & PFN v2 & PFN v2.5 & PFN v2.6 & PFN v3 & ICL v1 & ICL v2 \\
\midrule
\endfirsthead
\toprule
Condition & PFN v1 & PFN v2 & PFN v2.5 & PFN v2.6 & PFN v3 & ICL v1 & ICL v2 \\
\midrule
\endhead
\midrule
\multicolumn{8}{r}{\scriptsize Continued on next page} \\
\endfoot
\bottomrule
\endlastfoot
Identity & $0.7680 \mathbin{\pm} 0.0244$ & $0.7638 \mathbin{\pm} 0.0233$ & $0.7676 \mathbin{\pm} 0.0228$ & $0.7633 \mathbin{\pm} 0.0235$ & $0.7645 \mathbin{\pm} 0.0240$ & $0.7682 \mathbin{\pm} 0.0236$ & $0.7658 \mathbin{\pm} 0.0246$ \\
\addlinespace
Standard scaling & $0.7678 \mathbin{\pm} 0.0246$ & $0.7652 \mathbin{\pm} 0.0246$ & $0.7673 \mathbin{\pm} 0.0251$ & $0.7634 \mathbin{\pm} 0.0233$ & $0.7652 \mathbin{\pm} 0.0242$ & $0.7681 \mathbin{\pm} 0.0236$ & $0.7658 \mathbin{\pm} 0.0246$ \\
Log transform & $0.7635 \mathbin{\pm} 0.0267$ & $0.7701 \mathbin{\pm} 0.0237$ & $0.7665 \mathbin{\pm} 0.0231$ & $0.7643 \mathbin{\pm} 0.0220$ & $0.7632 \mathbin{\pm} 0.0254$ & $0.7689 \mathbin{\pm} 0.0252$ & $0.7661 \mathbin{\pm} 0.0245$ \\
Square-root transform & $0.7643 \mathbin{\pm} 0.0261$ & $0.7678 \mathbin{\pm} 0.0221$ & $0.7658 \mathbin{\pm} 0.0238$ & $0.7635 \mathbin{\pm} 0.0226$ & $0.7661 \mathbin{\pm} 0.0239$ & $0.7673 \mathbin{\pm} 0.0259$ & $0.7652 \mathbin{\pm} 0.0247$ \\
Box–Cox & $0.7685 \mathbin{\pm} 0.0256$ & $0.7668 \mathbin{\pm} 0.0231$ & $0.7656 \mathbin{\pm} 0.0234$ & $0.7634 \mathbin{\pm} 0.0235$ & $0.7654 \mathbin{\pm} 0.0236$ & $0.7656 \mathbin{\pm} 0.0248$ & $0.7652 \mathbin{\pm} 0.0249$ \\
Min–max scaling & $0.7678 \mathbin{\pm} 0.0246$ & $0.7652 \mathbin{\pm} 0.0234$ & $0.7671 \mathbin{\pm} 0.0247$ & $0.7630 \mathbin{\pm} 0.0234$ & $0.7648 \mathbin{\pm} 0.0249$ & $0.7681 \mathbin{\pm} 0.0236$ & $0.7658 \mathbin{\pm} 0.0246$ \\
\addlinespace
One-hot encoding & --- & --- & --- & --- & --- & --- & --- \\
Ordinal encoding & --- & --- & --- & --- & --- & --- & --- \\
\end{longtable}
\endgroup

\begingroup
\scriptsize
\setlength{\tabcolsep}{0.65mm}
\begin{longtable}{@{}lccccccc@{}}
\caption{Complete TabPFN and TabICL raw metric results for Diabetes, metric macro f1. Values are mean $\mathbin{\pm}$ sample standard deviation across the 30 paired splits.}\label{tab:supp-raw-21}\label{tab:supp-raw-22} \\
\toprule
Condition & PFN v1 & PFN v2 & PFN v2.5 & PFN v2.6 & PFN v3 & ICL v1 & ICL v2 \\
\midrule
\endfirsthead
\toprule
Condition & PFN v1 & PFN v2 & PFN v2.5 & PFN v2.6 & PFN v3 & ICL v1 & ICL v2 \\
\midrule
\endhead
\midrule
\multicolumn{8}{r}{\scriptsize Continued on next page} \\
\endfoot
\bottomrule
\endlastfoot
Identity & $0.7331 \mathbin{\pm} 0.0268$ & $0.7299 \mathbin{\pm} 0.0257$ & $0.7337 \mathbin{\pm} 0.0249$ & $0.7293 \mathbin{\pm} 0.0265$ & $0.7301 \mathbin{\pm} 0.0263$ & $0.7333 \mathbin{\pm} 0.0252$ & $0.7303 \mathbin{\pm} 0.0278$ \\
\addlinespace
Standard scaling & $0.7330 \mathbin{\pm} 0.0269$ & $0.7312 \mathbin{\pm} 0.0271$ & $0.7333 \mathbin{\pm} 0.0276$ & $0.7295 \mathbin{\pm} 0.0261$ & $0.7313 \mathbin{\pm} 0.0266$ & $0.7332 \mathbin{\pm} 0.0253$ & $0.7303 \mathbin{\pm} 0.0278$ \\
Log transform & $0.7285 \mathbin{\pm} 0.0294$ & $0.7381 \mathbin{\pm} 0.0259$ & $0.7333 \mathbin{\pm} 0.0247$ & $0.7304 \mathbin{\pm} 0.0245$ & $0.7300 \mathbin{\pm} 0.0280$ & $0.7345 \mathbin{\pm} 0.0284$ & $0.7308 \mathbin{\pm} 0.0279$ \\
Square-root transform & $0.7288 \mathbin{\pm} 0.0287$ & $0.7350 \mathbin{\pm} 0.0242$ & $0.7320 \mathbin{\pm} 0.0263$ & $0.7296 \mathbin{\pm} 0.0249$ & $0.7327 \mathbin{\pm} 0.0264$ & $0.7322 \mathbin{\pm} 0.0286$ & $0.7295 \mathbin{\pm} 0.0280$ \\
Box–Cox & $0.7346 \mathbin{\pm} 0.0281$ & $0.7326 \mathbin{\pm} 0.0259$ & $0.7309 \mathbin{\pm} 0.0257$ & $0.7293 \mathbin{\pm} 0.0262$ & $0.7313 \mathbin{\pm} 0.0260$ & $0.7301 \mathbin{\pm} 0.0278$ & $0.7300 \mathbin{\pm} 0.0283$ \\
Min–max scaling & $0.7330 \mathbin{\pm} 0.0269$ & $0.7315 \mathbin{\pm} 0.0258$ & $0.7328 \mathbin{\pm} 0.0266$ & $0.7290 \mathbin{\pm} 0.0264$ & $0.7308 \mathbin{\pm} 0.0275$ & $0.7332 \mathbin{\pm} 0.0253$ & $0.7303 \mathbin{\pm} 0.0278$ \\
\addlinespace
One-hot encoding & --- & --- & --- & --- & --- & --- & --- \\
Ordinal encoding & --- & --- & --- & --- & --- & --- & --- \\
\end{longtable}
\endgroup

\begingroup
\scriptsize
\setlength{\tabcolsep}{0.65mm}
\begin{longtable}{@{}lccccccc@{}}
\caption{Complete TabPFN and TabICL raw metric results for Fitness club, metric accuracy. Values are mean $\mathbin{\pm}$ sample standard deviation across the 30 paired splits.}\label{tab:supp-raw-23}\label{tab:supp-raw-24} \\
\toprule
Condition & PFN v1 & PFN v2 & PFN v2.5 & PFN v2.6 & PFN v3 & ICL v1 & ICL v2 \\
\midrule
\endfirsthead
\toprule
Condition & PFN v1 & PFN v2 & PFN v2.5 & PFN v2.6 & PFN v3 & ICL v1 & ICL v2 \\
\midrule
\endhead
\midrule
\multicolumn{8}{r}{\scriptsize Continued on next page} \\
\endfoot
\bottomrule
\endlastfoot
Identity & $0.7776 \mathbin{\pm} 0.0139$ & $0.7809 \mathbin{\pm} 0.0108$ & $0.7791 \mathbin{\pm} 0.0109$ & $0.7805 \mathbin{\pm} 0.0110$ & $0.7815 \mathbin{\pm} 0.0107$ & $0.7774 \mathbin{\pm} 0.0119$ & $0.7817 \mathbin{\pm} 0.0113$ \\
\addlinespace
Standard scaling & $0.7774 \mathbin{\pm} 0.0135$ & $0.7815 \mathbin{\pm} 0.0104$ & $0.7799 \mathbin{\pm} 0.0109$ & $0.7805 \mathbin{\pm} 0.0109$ & $0.7811 \mathbin{\pm} 0.0108$ & $0.7773 \mathbin{\pm} 0.0120$ & $0.7817 \mathbin{\pm} 0.0113$ \\
Log transform & $0.7771 \mathbin{\pm} 0.0144$ & $0.7819 \mathbin{\pm} 0.0112$ & $0.7807 \mathbin{\pm} 0.0117$ & $0.7797 \mathbin{\pm} 0.0111$ & $0.7810 \mathbin{\pm} 0.0113$ & $0.7777 \mathbin{\pm} 0.0125$ & $0.7817 \mathbin{\pm} 0.0114$ \\
Square-root transform & $0.7769 \mathbin{\pm} 0.0140$ & $0.7814 \mathbin{\pm} 0.0110$ & $0.7803 \mathbin{\pm} 0.0114$ & $0.7805 \mathbin{\pm} 0.0113$ & $0.7818 \mathbin{\pm} 0.0119$ & $0.7777 \mathbin{\pm} 0.0126$ & $0.7815 \mathbin{\pm} 0.0111$ \\
Box–Cox & $0.7773 \mathbin{\pm} 0.0141$ & $0.7813 \mathbin{\pm} 0.0112$ & $0.7808 \mathbin{\pm} 0.0126$ & $0.7811 \mathbin{\pm} 0.0107$ & $0.7817 \mathbin{\pm} 0.0106$ & $0.7773 \mathbin{\pm} 0.0128$ & $0.7817 \mathbin{\pm} 0.0114$ \\
Min–max scaling & $0.7774 \mathbin{\pm} 0.0135$ & $0.7814 \mathbin{\pm} 0.0106$ & $0.7797 \mathbin{\pm} 0.0114$ & $0.7804 \mathbin{\pm} 0.0109$ & $0.7814 \mathbin{\pm} 0.0107$ & $0.7774 \mathbin{\pm} 0.0121$ & $0.7817 \mathbin{\pm} 0.0113$ \\
\addlinespace
One-hot encoding & $0.7720 \mathbin{\pm} 0.0149$ & $0.7808 \mathbin{\pm} 0.0112$ & $0.7797 \mathbin{\pm} 0.0107$ & $0.7809 \mathbin{\pm} 0.0106$ & $0.7807 \mathbin{\pm} 0.0106$ & $0.7777 \mathbin{\pm} 0.0128$ & $0.7815 \mathbin{\pm} 0.0110$ \\
Ordinal encoding & $0.7776 \mathbin{\pm} 0.0139$ & $0.7811 \mathbin{\pm} 0.0110$ & $0.7803 \mathbin{\pm} 0.0115$ & $0.7804 \mathbin{\pm} 0.0110$ & $0.7808 \mathbin{\pm} 0.0114$ & $0.7774 \mathbin{\pm} 0.0119$ & $0.7819 \mathbin{\pm} 0.0111$ \\
\end{longtable}
\endgroup

\begingroup
\scriptsize
\setlength{\tabcolsep}{0.65mm}
\begin{longtable}{@{}lccccccc@{}}
\caption{Complete TabPFN and TabICL raw metric results for Fitness club, metric macro f1. Values are mean $\mathbin{\pm}$ sample standard deviation across the 30 paired splits.}\label{tab:supp-raw-25}\label{tab:supp-raw-26} \\
\toprule
Condition & PFN v1 & PFN v2 & PFN v2.5 & PFN v2.6 & PFN v3 & ICL v1 & ICL v2 \\
\midrule
\endfirsthead
\toprule
Condition & PFN v1 & PFN v2 & PFN v2.5 & PFN v2.6 & PFN v3 & ICL v1 & ICL v2 \\
\midrule
\endhead
\midrule
\multicolumn{8}{r}{\scriptsize Continued on next page} \\
\endfoot
\bottomrule
\endlastfoot
Identity & $0.7104 \mathbin{\pm} 0.0211$ & $0.7162 \mathbin{\pm} 0.0178$ & $0.7141 \mathbin{\pm} 0.0188$ & $0.7181 \mathbin{\pm} 0.0181$ & $0.7177 \mathbin{\pm} 0.0175$ & $0.7115 \mathbin{\pm} 0.0190$ & $0.7156 \mathbin{\pm} 0.0190$ \\
\addlinespace
Standard scaling & $0.7102 \mathbin{\pm} 0.0207$ & $0.7173 \mathbin{\pm} 0.0171$ & $0.7154 \mathbin{\pm} 0.0190$ & $0.7182 \mathbin{\pm} 0.0180$ & $0.7169 \mathbin{\pm} 0.0177$ & $0.7117 \mathbin{\pm} 0.0189$ & $0.7156 \mathbin{\pm} 0.0190$ \\
Log transform & $0.7091 \mathbin{\pm} 0.0219$ & $0.7158 \mathbin{\pm} 0.0178$ & $0.7157 \mathbin{\pm} 0.0187$ & $0.7170 \mathbin{\pm} 0.0175$ & $0.7169 \mathbin{\pm} 0.0183$ & $0.7112 \mathbin{\pm} 0.0194$ & $0.7156 \mathbin{\pm} 0.0187$ \\
Square-root transform & $0.7085 \mathbin{\pm} 0.0215$ & $0.7153 \mathbin{\pm} 0.0178$ & $0.7142 \mathbin{\pm} 0.0192$ & $0.7180 \mathbin{\pm} 0.0184$ & $0.7171 \mathbin{\pm} 0.0191$ & $0.7111 \mathbin{\pm} 0.0195$ & $0.7151 \mathbin{\pm} 0.0186$ \\
Box–Cox & $0.7095 \mathbin{\pm} 0.0219$ & $0.7153 \mathbin{\pm} 0.0183$ & $0.7165 \mathbin{\pm} 0.0195$ & $0.7186 \mathbin{\pm} 0.0176$ & $0.7180 \mathbin{\pm} 0.0169$ & $0.7106 \mathbin{\pm} 0.0199$ & $0.7156 \mathbin{\pm} 0.0188$ \\
Min–max scaling & $0.7102 \mathbin{\pm} 0.0207$ & $0.7166 \mathbin{\pm} 0.0177$ & $0.7148 \mathbin{\pm} 0.0192$ & $0.7180 \mathbin{\pm} 0.0180$ & $0.7173 \mathbin{\pm} 0.0174$ & $0.7118 \mathbin{\pm} 0.0190$ & $0.7156 \mathbin{\pm} 0.0190$ \\
\addlinespace
One-hot encoding & $0.7049 \mathbin{\pm} 0.0210$ & $0.7157 \mathbin{\pm} 0.0185$ & $0.7155 \mathbin{\pm} 0.0177$ & $0.7190 \mathbin{\pm} 0.0172$ & $0.7166 \mathbin{\pm} 0.0164$ & $0.7144 \mathbin{\pm} 0.0209$ & $0.7156 \mathbin{\pm} 0.0185$ \\
Ordinal encoding & $0.7104 \mathbin{\pm} 0.0211$ & $0.7168 \mathbin{\pm} 0.0175$ & $0.7163 \mathbin{\pm} 0.0183$ & $0.7178 \mathbin{\pm} 0.0180$ & $0.7170 \mathbin{\pm} 0.0181$ & $0.7115 \mathbin{\pm} 0.0190$ & $0.7160 \mathbin{\pm} 0.0192$ \\
\end{longtable}
\endgroup

\begingroup
\scriptsize
\setlength{\tabcolsep}{1.35mm}
\begin{longtable}{@{}lccccc@{}}
\caption{Complete raw metric results for Healthcare insurance, metric r2. Values are mean $\mathbin{\pm}$ sample standard deviation across the 30 paired splits.}\label{tab:supp-raw-27} \\
\toprule
Condition & PFN v2 & PFN v2.5 & PFN v2.6 & PFN v3 & ICL v2 \\
\midrule
\endfirsthead
\toprule
Condition & PFN v2 & PFN v2.5 & PFN v2.6 & PFN v3 & ICL v2 \\
\midrule
\endhead
\midrule
\multicolumn{6}{r}{\scriptsize Continued on next page} \\
\endfoot
\bottomrule
\endlastfoot
Identity & $0.8479 \mathbin{\pm} 0.0182$ & $0.8598 \mathbin{\pm} 0.0196$ & $0.8650 \mathbin{\pm} 0.0192$ & $0.8634 \mathbin{\pm} 0.0201$ & $0.8636 \mathbin{\pm} 0.0180$ \\
\addlinespace
Standard scaling & $0.8475 \mathbin{\pm} 0.0183$ & $0.8600 \mathbin{\pm} 0.0196$ & $0.8626 \mathbin{\pm} 0.0192$ & $0.8633 \mathbin{\pm} 0.0201$ & $0.8636 \mathbin{\pm} 0.0180$ \\
Log transform & $0.8478 \mathbin{\pm} 0.0186$ & $0.8599 \mathbin{\pm} 0.0195$ & $0.8651 \mathbin{\pm} 0.0191$ & $0.8629 \mathbin{\pm} 0.0199$ & $0.8637 \mathbin{\pm} 0.0181$ \\
Square-root transform & $0.8472 \mathbin{\pm} 0.0188$ & $0.8598 \mathbin{\pm} 0.0196$ & $0.8650 \mathbin{\pm} 0.0191$ & $0.8632 \mathbin{\pm} 0.0200$ & $0.8636 \mathbin{\pm} 0.0180$ \\
Box–Cox & $0.8479 \mathbin{\pm} 0.0185$ & $0.8598 \mathbin{\pm} 0.0198$ & $0.8653 \mathbin{\pm} 0.0194$ & $0.8634 \mathbin{\pm} 0.0200$ & $0.8635 \mathbin{\pm} 0.0181$ \\
Min–max scaling & $0.8479 \mathbin{\pm} 0.0183$ & $0.8600 \mathbin{\pm} 0.0196$ & $0.8648 \mathbin{\pm} 0.0190$ & $0.8633 \mathbin{\pm} 0.0200$ & $0.8636 \mathbin{\pm} 0.0180$ \\
\addlinespace
One-hot encoding & $0.8441 \mathbin{\pm} 0.0178$ & $0.8612 \mathbin{\pm} 0.0201$ & $0.8648 \mathbin{\pm} 0.0193$ & $0.8627 \mathbin{\pm} 0.0200$ & $0.8635 \mathbin{\pm} 0.0182$ \\
Ordinal encoding & $0.8478 \mathbin{\pm} 0.0185$ & $0.8596 \mathbin{\pm} 0.0201$ & $0.8649 \mathbin{\pm} 0.0191$ & $0.8633 \mathbin{\pm} 0.0200$ & $0.8634 \mathbin{\pm} 0.0180$ \\
\end{longtable}
\endgroup

\begingroup
\scriptsize
\setlength{\tabcolsep}{1.35mm}
\begin{longtable}{@{}lccccc@{}}
\caption{Complete raw metric results for Healthcare insurance, metric rmse. Values are mean $\mathbin{\pm}$ sample standard deviation across the 30 paired splits.}\label{tab:supp-raw-28} \\
\toprule
Condition & PFN v2 & PFN v2.5 & PFN v2.6 & PFN v3 & ICL v2 \\
\midrule
\endfirsthead
\toprule
Condition & PFN v2 & PFN v2.5 & PFN v2.6 & PFN v3 & ICL v2 \\
\midrule
\endhead
\midrule
\multicolumn{6}{r}{\scriptsize Continued on next page} \\
\endfoot
\bottomrule
\endlastfoot
Identity & $4704.2 \mathbin{\pm} 301.9$ & $4513.1 \mathbin{\pm} 334.5$ & $4428.3 \mathbin{\pm} 334.2$ & $4453.9 \mathbin{\pm} 346.1$ & $4452.3 \mathbin{\pm} 310.6$ \\
\addlinespace
Standard scaling & $4710.4 \mathbin{\pm} 304.8$ & $4509.6 \mathbin{\pm} 335.4$ & $4468.2 \mathbin{\pm} 338.8$ & $4455.7 \mathbin{\pm} 345.7$ & $4452.3 \mathbin{\pm} 310.6$ \\
Log transform & $4705.2 \mathbin{\pm} 305.5$ & $4511.7 \mathbin{\pm} 333.6$ & $4427.5 \mathbin{\pm} 331.7$ & $4461.3 \mathbin{\pm} 343.8$ & $4450.5 \mathbin{\pm} 310.5$ \\
Square-root transform & $4714.0 \mathbin{\pm} 306.9$ & $4512.9 \mathbin{\pm} 334.5$ & $4428.6 \mathbin{\pm} 331.2$ & $4457.2 \mathbin{\pm} 344.0$ & $4452.3 \mathbin{\pm} 310.6$ \\
Box–Cox & $4703.0 \mathbin{\pm} 303.4$ & $4512.5 \mathbin{\pm} 340.7$ & $4422.8 \mathbin{\pm} 335.7$ & $4453.5 \mathbin{\pm} 343.3$ & $4453.7 \mathbin{\pm} 309.0$ \\
Min–max scaling & $4703.5 \mathbin{\pm} 302.3$ & $4510.4 \mathbin{\pm} 334.5$ & $4431.8 \mathbin{\pm} 330.3$ & $4454.6 \mathbin{\pm} 345.0$ & $4452.3 \mathbin{\pm} 310.6$ \\
\addlinespace
One-hot encoding & $4762.3 \mathbin{\pm} 292.5$ & $4489.8 \mathbin{\pm} 345.5$ & $4430.6 \mathbin{\pm} 333.0$ & $4464.4 \mathbin{\pm} 342.7$ & $4453.7 \mathbin{\pm} 313.0$ \\
Ordinal encoding & $4705.3 \mathbin{\pm} 307.1$ & $4516.7 \mathbin{\pm} 344.0$ & $4430.5 \mathbin{\pm} 331.9$ & $4455.9 \mathbin{\pm} 346.2$ & $4455.9 \mathbin{\pm} 311.6$ \\
\end{longtable}
\endgroup

\begingroup
\scriptsize
\setlength{\tabcolsep}{0.65mm}
\begin{longtable}{@{}lccccccc@{}}
\caption{Complete TabPFN and TabICL raw metric results for Maternal health, metric accuracy. Values are mean $\mathbin{\pm}$ sample standard deviation across the 30 paired splits.}\label{tab:supp-raw-29}\label{tab:supp-raw-30} \\
\toprule
Condition & PFN v1 & PFN v2 & PFN v2.5 & PFN v2.6 & PFN v3 & ICL v1 & ICL v2 \\
\midrule
\endfirsthead
\toprule
Condition & PFN v1 & PFN v2 & PFN v2.5 & PFN v2.6 & PFN v3 & ICL v1 & ICL v2 \\
\midrule
\endhead
\midrule
\multicolumn{8}{r}{\scriptsize Continued on next page} \\
\endfoot
\bottomrule
\endlastfoot
Identity & $0.7739 \mathbin{\pm} 0.0257$ & $0.8370 \mathbin{\pm} 0.0234$ & $0.8321 \mathbin{\pm} 0.0223$ & $0.8391 \mathbin{\pm} 0.0248$ & $0.8411 \mathbin{\pm} 0.0238$ & $0.8290 \mathbin{\pm} 0.0218$ & $0.8451 \mathbin{\pm} 0.0266$ \\
\addlinespace
Standard scaling & $0.7739 \mathbin{\pm} 0.0257$ & $0.8362 \mathbin{\pm} 0.0225$ & $0.8319 \mathbin{\pm} 0.0212$ & $0.8390 \mathbin{\pm} 0.0248$ & $0.8410 \mathbin{\pm} 0.0233$ & $0.8290 \mathbin{\pm} 0.0218$ & $0.8451 \mathbin{\pm} 0.0266$ \\
Log transform & $0.7741 \mathbin{\pm} 0.0276$ & $0.8374 \mathbin{\pm} 0.0250$ & $0.8329 \mathbin{\pm} 0.0224$ & $0.8395 \mathbin{\pm} 0.0223$ & $0.8427 \mathbin{\pm} 0.0237$ & $0.8299 \mathbin{\pm} 0.0227$ & $0.8437 \mathbin{\pm} 0.0261$ \\
Square-root transform & $0.7739 \mathbin{\pm} 0.0263$ & $0.8372 \mathbin{\pm} 0.0221$ & $0.8323 \mathbin{\pm} 0.0225$ & $0.8388 \mathbin{\pm} 0.0239$ & $0.8421 \mathbin{\pm} 0.0232$ & $0.8297 \mathbin{\pm} 0.0223$ & $0.8444 \mathbin{\pm} 0.0257$ \\
Box–Cox & $0.7735 \mathbin{\pm} 0.0258$ & $0.8386 \mathbin{\pm} 0.0247$ & $0.8368 \mathbin{\pm} 0.0231$ & $0.8380 \mathbin{\pm} 0.0227$ & $0.8402 \mathbin{\pm} 0.0228$ & $0.8293 \mathbin{\pm} 0.0233$ & $0.8433 \mathbin{\pm} 0.0250$ \\
Min–max scaling & $0.7739 \mathbin{\pm} 0.0257$ & $0.8375 \mathbin{\pm} 0.0229$ & $0.8331 \mathbin{\pm} 0.0225$ & $0.8391 \mathbin{\pm} 0.0248$ & $0.8412 \mathbin{\pm} 0.0221$ & $0.8290 \mathbin{\pm} 0.0218$ & $0.8451 \mathbin{\pm} 0.0266$ \\
\addlinespace
One-hot encoding & --- & --- & --- & --- & --- & --- & --- \\
Ordinal encoding & --- & --- & --- & --- & --- & --- & --- \\
\end{longtable}
\endgroup

\begingroup
\scriptsize
\setlength{\tabcolsep}{0.65mm}
\begin{longtable}{@{}lccccccc@{}}
\caption{Complete TabPFN and TabICL raw metric results for Maternal health, metric macro f1. Values are mean $\mathbin{\pm}$ sample standard deviation across the 30 paired splits.}\label{tab:supp-raw-31}\label{tab:supp-raw-32} \\
\toprule
Condition & PFN v1 & PFN v2 & PFN v2.5 & PFN v2.6 & PFN v3 & ICL v1 & ICL v2 \\
\midrule
\endfirsthead
\toprule
Condition & PFN v1 & PFN v2 & PFN v2.5 & PFN v2.6 & PFN v3 & ICL v1 & ICL v2 \\
\midrule
\endhead
\midrule
\multicolumn{8}{r}{\scriptsize Continued on next page} \\
\endfoot
\bottomrule
\endlastfoot
Identity & $0.7760 \mathbin{\pm} 0.0262$ & $0.8390 \mathbin{\pm} 0.0235$ & $0.8344 \mathbin{\pm} 0.0223$ & $0.8412 \mathbin{\pm} 0.0249$ & $0.8433 \mathbin{\pm} 0.0237$ & $0.8314 \mathbin{\pm} 0.0224$ & $0.8471 \mathbin{\pm} 0.0265$ \\
\addlinespace
Standard scaling & $0.7760 \mathbin{\pm} 0.0262$ & $0.8382 \mathbin{\pm} 0.0224$ & $0.8341 \mathbin{\pm} 0.0214$ & $0.8411 \mathbin{\pm} 0.0249$ & $0.8434 \mathbin{\pm} 0.0232$ & $0.8314 \mathbin{\pm} 0.0224$ & $0.8471 \mathbin{\pm} 0.0265$ \\
Log transform & $0.7759 \mathbin{\pm} 0.0277$ & $0.8396 \mathbin{\pm} 0.0248$ & $0.8351 \mathbin{\pm} 0.0224$ & $0.8419 \mathbin{\pm} 0.0224$ & $0.8449 \mathbin{\pm} 0.0237$ & $0.8323 \mathbin{\pm} 0.0232$ & $0.8457 \mathbin{\pm} 0.0262$ \\
Square-root transform & $0.7758 \mathbin{\pm} 0.0267$ & $0.8392 \mathbin{\pm} 0.0221$ & $0.8347 \mathbin{\pm} 0.0223$ & $0.8411 \mathbin{\pm} 0.0240$ & $0.8445 \mathbin{\pm} 0.0230$ & $0.8321 \mathbin{\pm} 0.0230$ & $0.8464 \mathbin{\pm} 0.0258$ \\
Box–Cox & $0.7755 \mathbin{\pm} 0.0255$ & $0.8405 \mathbin{\pm} 0.0244$ & $0.8385 \mathbin{\pm} 0.0233$ & $0.8402 \mathbin{\pm} 0.0228$ & $0.8428 \mathbin{\pm} 0.0226$ & $0.8318 \mathbin{\pm} 0.0237$ & $0.8454 \mathbin{\pm} 0.0250$ \\
Min–max scaling & $0.7760 \mathbin{\pm} 0.0262$ & $0.8395 \mathbin{\pm} 0.0225$ & $0.8355 \mathbin{\pm} 0.0224$ & $0.8412 \mathbin{\pm} 0.0249$ & $0.8436 \mathbin{\pm} 0.0223$ & $0.8314 \mathbin{\pm} 0.0224$ & $0.8471 \mathbin{\pm} 0.0265$ \\
\addlinespace
One-hot encoding & --- & --- & --- & --- & --- & --- & --- \\
Ordinal encoding & --- & --- & --- & --- & --- & --- & --- \\
\end{longtable}
\endgroup

\begingroup
\scriptsize
\setlength{\tabcolsep}{0.65mm}
\begin{longtable}{@{}lccccccc@{}}
\caption{Complete TabPFN and TabICL raw metric results for QSAR biodeg, metric accuracy. Values are mean $\mathbin{\pm}$ sample standard deviation across the 30 paired splits.}\label{tab:supp-raw-33}\label{tab:supp-raw-34} \\
\toprule
Condition & PFN v1 & PFN v2 & PFN v2.5 & PFN v2.6 & PFN v3 & ICL v1 & ICL v2 \\
\midrule
\endfirsthead
\toprule
Condition & PFN v1 & PFN v2 & PFN v2.5 & PFN v2.6 & PFN v3 & ICL v1 & ICL v2 \\
\midrule
\endhead
\midrule
\multicolumn{8}{r}{\scriptsize Continued on next page} \\
\endfoot
\bottomrule
\endlastfoot
Identity & $0.8839 \mathbin{\pm} 0.0162$ & $0.8766 \mathbin{\pm} 0.0147$ & $0.8782 \mathbin{\pm} 0.0166$ & $0.8749 \mathbin{\pm} 0.0166$ & $0.8822 \mathbin{\pm} 0.0148$ & $0.8780 \mathbin{\pm} 0.0171$ & $0.8845 \mathbin{\pm} 0.0134$ \\
\addlinespace
Standard scaling & $0.8833 \mathbin{\pm} 0.0161$ & $0.8770 \mathbin{\pm} 0.0151$ & $0.8782 \mathbin{\pm} 0.0158$ & $0.8759 \mathbin{\pm} 0.0163$ & $0.8807 \mathbin{\pm} 0.0129$ & $0.8784 \mathbin{\pm} 0.0175$ & $0.8845 \mathbin{\pm} 0.0134$ \\
Log transform & $0.8809 \mathbin{\pm} 0.0147$ & $0.8756 \mathbin{\pm} 0.0147$ & $0.8787 \mathbin{\pm} 0.0152$ & $0.8760 \mathbin{\pm} 0.0163$ & $0.8813 \mathbin{\pm} 0.0142$ & $0.8744 \mathbin{\pm} 0.0171$ & $0.8839 \mathbin{\pm} 0.0142$ \\
Square-root transform & $0.8767 \mathbin{\pm} 0.0165$ & $0.8767 \mathbin{\pm} 0.0152$ & $0.8785 \mathbin{\pm} 0.0157$ & $0.8747 \mathbin{\pm} 0.0166$ & $0.8808 \mathbin{\pm} 0.0135$ & $0.8726 \mathbin{\pm} 0.0176$ & $0.8842 \mathbin{\pm} 0.0137$ \\
Box–Cox & $0.8817 \mathbin{\pm} 0.0161$ & $0.8768 \mathbin{\pm} 0.0160$ & $0.8783 \mathbin{\pm} 0.0158$ & $0.8745 \mathbin{\pm} 0.0170$ & $0.8825 \mathbin{\pm} 0.0149$ & $0.8784 \mathbin{\pm} 0.0166$ & $0.8834 \mathbin{\pm} 0.0129$ \\
Min–max scaling & $0.8833 \mathbin{\pm} 0.0161$ & $0.8760 \mathbin{\pm} 0.0156$ & $0.8780 \mathbin{\pm} 0.0154$ & $0.8752 \mathbin{\pm} 0.0165$ & $0.8808 \mathbin{\pm} 0.0126$ & $0.8782 \mathbin{\pm} 0.0175$ & $0.8845 \mathbin{\pm} 0.0134$ \\
\addlinespace
One-hot encoding & $0.8789 \mathbin{\pm} 0.0164$ & $0.8756 \mathbin{\pm} 0.0151$ & $0.8787 \mathbin{\pm} 0.0157$ & $0.8768 \mathbin{\pm} 0.0140$ & $0.8806 \mathbin{\pm} 0.0150$ & $0.8781 \mathbin{\pm} 0.0181$ & $0.8847 \mathbin{\pm} 0.0126$ \\
Ordinal encoding & $0.8847 \mathbin{\pm} 0.0160$ & $0.8753 \mathbin{\pm} 0.0156$ & $0.8775 \mathbin{\pm} 0.0164$ & $0.8734 \mathbin{\pm} 0.0158$ & $0.8818 \mathbin{\pm} 0.0158$ & $0.8771 \mathbin{\pm} 0.0181$ & $0.8846 \mathbin{\pm} 0.0153$ \\
\end{longtable}%
\endgroup

\end{document}